\documentclass{article}

\usepackage{microtype}
\usepackage{graphicx}
\usepackage{subcaption}
\usepackage{booktabs} % for professional tables

\usepackage{hyperref}

\usepackage[preprint]{icml2026}

\usepackage{amsmath}
\usepackage{amssymb}
\usepackage{mathtools}
\usepackage{amsthm}

\usepackage[capitalize,noabbrev]{cleveref}

\theoremstyle{plain}

\theoremstyle{definition}

\theoremstyle{remark}

\usepackage[disable,textsize=tiny]{todonotes}
\icmltitlerunning{Universal Hypernetworks for Arbitrary Models}

\begin{document}

\twocolumn[
  \icmltitle{Universal Hypernetworks for Arbitrary Models}



%   \icmlsetsymbol{equal}{*}

  \begin{icmlauthorlist}
    \icmlauthor{Xuanfeng Zhou}{ind}
  \end{icmlauthorlist}

    \icmlaffiliation{ind}{Independent Researcher}

  \icmlcorrespondingauthor{Xuanfeng Zhou}{xuanfeng.zhou.research@gmail.com, zxf12138@buaa.edu.cn}

  \icmlkeywords{Hypernetworks, Implicit Neural Representations, Neural Network Parameterization}

  \vskip 0.3in
]



\printAffiliationsAndNotice{}  % no special notice (required even if empty)

%%%%%%%%%%%%%%%%%%%%%%%%%%%%%%%%
% Abstract
%%%%%%%%%%%%%%%%%%%%%%%%%%%%%%%%
\begin{abstract}
Conventional hypernetworks are typically engineered around a specific base-model parameterization, so changing the target architecture often entails redesigning the hypernetwork and retraining it from scratch.
We introduce the \emph{Universal Hypernetwork} (UHN), a fixed-architecture generator that predicts weights from deterministic parameter, architecture, and task descriptors.
This descriptor-based formulation decouples the generator architecture from target-network parameterization, so one generator can instantiate heterogeneous models across the tested architecture and task families.
Our empirical claims are threefold: (1) one fixed UHN remains competitive with direct training across vision, graph, text, and formula-regression benchmarks;
(2) the same UHN supports both multi-model generalization within a family and multi-task learning across heterogeneous models; and
(3) UHN enables stable recursive generation with up to three intermediate generated UHNs before the final base model.
Our code is available at \url{https://github.com/Xuanfeng-Zhou/UHN}.
\end{abstract}

%%%%%%%%%%%%%%%%%%%%%%%%%%%%%%%%
% introduction
%%%%%%%%%%%%%%%%%%%%%%%%%%%%%%%%
\section{Introduction}
\label{introduction}
Hypernetworks generate neural-network parameters and are widely used for model compression \cite{ha2016hypernetworks}, neural architecture search \cite{zhang2018graph} and multi-task learning \cite{navon2020learning}.
However, many existing hypernetworks are tightly coupled to a specific target architecture and task: adapting to a new model or objective typically requires redesign and retraining \cite{chen2022transformers, navon2020learning}.
This architectural coupling can hinder scaling to heterogeneous model classes and limit use in multi-model and multi-task settings \cite{hedlin2025hypernet}.

We propose \emph{Universal Hypernetwork} (UHN), a single fixed-architecture generator that supports heterogeneous neural architectures, tasks, and modalities \emph{without changing the generator itself}.
Concretely, UHN predicts each scalar parameter from deterministic descriptors that encode parameter index, architecture information, and task information.

% Setting taxonomy
\begin{figure}[t]
    \centering
    \includegraphics[width=0.8\linewidth]{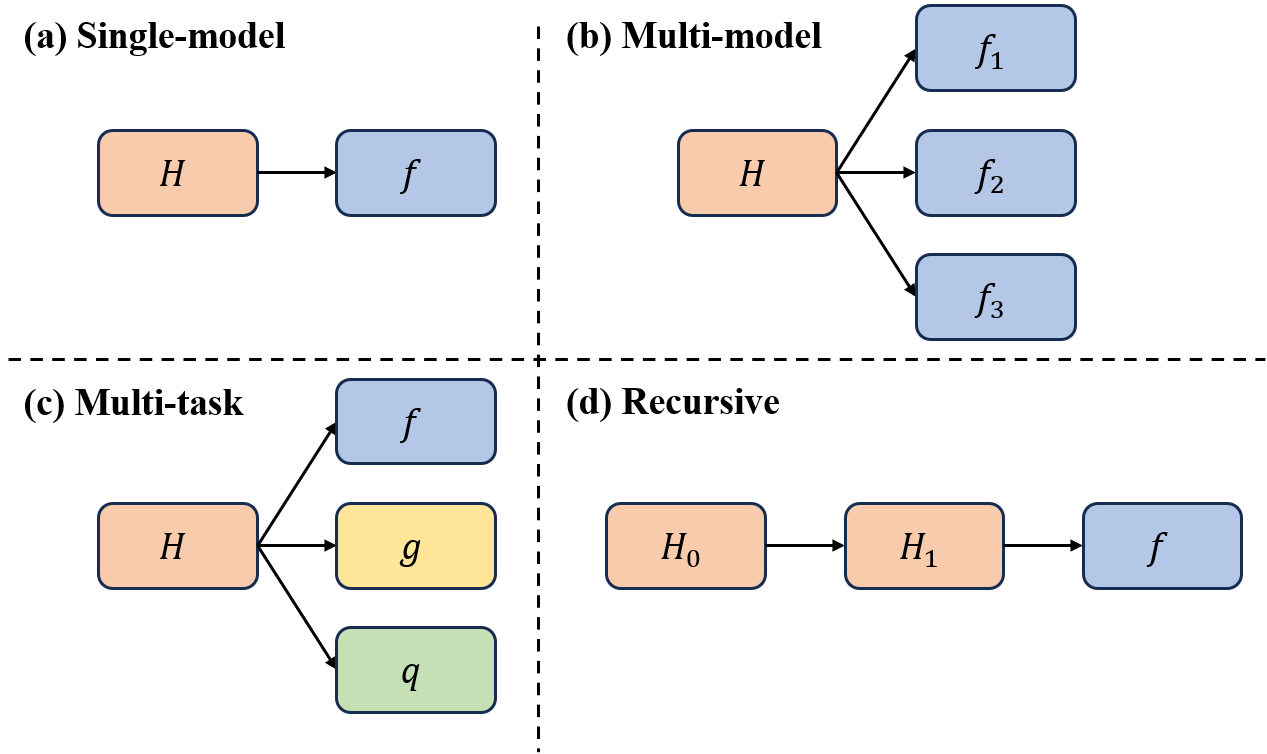}
    \caption{UHN settings: (a) single-model, UHN $H$ generates a single base model $f$; (b) multi-model, UHN $H$ generates a set of base models $\{f_i\}$ within one model family; (c) multi-task, UHN $H$ generates base models $f, g, q, \ldots$ across potentially heterogeneous architectures for different tasks; and (d) recursive generation, in which the root UHN $H_0$ generates an intermediate UHN $H_1$, which then generates the base model $f$.}
    \label{fig:setting-taxonomy}
\end{figure}

By moving model- and task-specificity into conditioning inputs (rather than architecture-specific output heads or learned per-block embedding sets \cite{hedlin2025hypernet, von2019continual}), UHN decouples hypernetwork design from target architecture and scales to heterogeneous model families without increasing generator complexity.

Our contributions are as follows:
\begin{itemize}
    \item \textbf{Single-generator versatility across settings.} With one fixed architecture, UHN generates heterogeneous target models and remains competitive with direct training across vision, graph, text, and formula-regression benchmarks.
    \item \textbf{Unified multi-model and multi-task learning.} The same UHN supports both model-family generalization (including held-out family members) and multi-task learning where tasks may use different architectures.
    \item \textbf{Stable recursive hypernetwork generation.} UHN can generate downstream UHNs and remains stable for up to three intermediate generated levels ($H_0 \rightarrow H_1 \rightarrow H_2 \rightarrow H_3 \rightarrow f$, where $H_0$ is the root UHN, each $H_i$ ($i\ge 1$) is generated by $H_{i-1}$, and the final base model $f$ is generated by $H_3$).
\end{itemize}

\Cref{fig:setting-taxonomy} summarizes these four settings used throughout the paper.
Empirically, UHN is competitive with direct training across diverse benchmarks (e.g., CIFAR-10 \cite{krizhevsky2009learning}, Cora \cite{sen2008collective}, and AG News \cite{zhang2015character}) with one shared generator architecture.
We further show that the same fixed UHN scales to larger and more diverse target networks while remaining effective for model family generalization, multi-task learning, and recursive generation.

%%%%%%%%%%%%%%%%%%%%%%%%%%%%%%%%
% Related Work
%%%%%%%%%%%%%%%%%%%%%%%%%%%%%%%%
\section{Related Work}
\label{related-work}
In this section, we focus on prior work most directly related to UHN's core claim: decoupling hypernetwork \cite{chauhan2024brief} design from target architecture and task.
Accordingly, we emphasize architecture-coupling limitations and descriptor-conditioned generation.
We use implicit neural representation (INR) \cite{essakine2024we} intuition as conceptual context because UHN models weights as a descriptor-to-value function, and we defer additional hypernetwork and INR discussion to \cref{app:additional-related-work}.

\subsection{Hypernetworks}
\label{related-hyper}
Hypernetworks are neural networks that generate the parameters of a target (``base'') network \cite{ha2016hypernetworks, chauhan2024brief}.
By parameterizing weights through a shared generator, hypernetworks can exploit structure and reuse across parameters, and have therefore been applied to model compression \cite{ha2016hypernetworks}, neural architecture search \cite{zhang2018graph}, multi-task learning \cite{navon2020learning}, and ensemble methods \cite{ratzlaff2019hypergan, deutsch2019generative}.

Despite these successes, many hypernetwork designs have limited flexibility and scalability: the hypernetwork output space is tightly coupled to the base-network architecture, so increasing the size (depth/width) of the base model typically requires a correspondingly larger and more complex hypernetwork \cite{chen2022transformers, navon2020learning}.
Several works mitigate this coupling by generating weights in blocks or components \cite{ha2016hypernetworks, von2019continual}, or by producing weights as a sequence using Transformers \cite{vaswani2017attention, chen2022transformers, ruiz2024hyperdreambooth, hedlin2025hypernet, gu2025foundation, kim2026nnit}.
However, these methods often rely on learned embeddings to represent blocks/components, and the number of such embeddings grows with the number of generated blocks \cite{hedlin2025hypernet, von2019continual}.
Consequently, adapting the hypernetwork to substantially different base architectures or tasks frequently still requires redesign and retraining \cite{ha2016hypernetworks}.

Several works use hypernetworks to generate \emph{multiple} base networks with varying architectures for purposes such as neural architecture search or parameter initialization by conditioning on an explicit representation of the target architecture (e.g., a memory-bank code or a graph-encoded description) \cite{brock2017smash, zhang2018graph, knyazev2021parameter, knyazev2023can, zhou2024logah, sukthanker2024multi, lv2025hypernas, kim2026nnit}.
A common limitation is that the hypernetwork is often implemented with an output space sized for the largest architecture in the family; when generating smaller models, this can lead to wasted outputs or ad-hoc masking/truncation \cite{zhou2024logah, kim2026nnit}.
Moreover, while these approaches can capture correlations between the performance of hypernetwork-generated models and fully trained models (supporting their use for search or initialization), the generated weights typically underperform direct end-to-end training and therefore require fine-tuning or retraining before deployment \cite{zhang2018graph, knyazev2023can}.

UHN parameterizes a base network by predicting each scalar weight from a deterministic description of its index (and, when applicable, structure and task descriptors), rather than from learned per-layer or per-block embeddings \cite{ha2016hypernetworks, von2019continual}.
We encode these descriptors using Gaussian Fourier features \cite{tancik2020fourier}, which provide a high-frequency basis for modeling complex weight fields.
This design allows a single fixed-architecture hypernetwork to generate parameters for different base architectures by changing only the conditioning inputs.

To generate model families, UHN conditions on an explicit architecture representation produced by a Transformer-based task-structure encoder \cite{vaswani2017attention}.
Unlike approaches whose output space is sized to the largest architecture in the family \cite{zhou2024logah, kim2026nnit}, this formulation does not require allocating parameters for absent layers/components when instantiating smaller models.
Conditioning additionally on task descriptors enables UHN to generate task-specific base models across heterogeneous modalities.

Finally, UHN supports recursive generation, i.e., generating the parameters of another UHN that then generates the base-model weights.
Recursive hypernetwork generation has been rarely explored and is challenging due to scaling and initialization instability \cite{liao2023generating, lutati2021hyperhypernetwork}.
Our formulation mitigates these issues by using explicit index/structure/task conditioning and an explicit initialization phase to improve optimization stability, enabling hierarchical (multi-level) model generation.

%%%%%%%%%%%%%%%%%%%%%%%%%%%%%%%%
% Method
%%%%%%%%%%%%%%%%%%%%%%%%%%%%%%%%
\section{Method}
\label{sec:method}
\subsection{Universal Hypernetwork}
\label{sec:uhn}

\subsubsection{Overview}
\begin{figure}[t]
    \centering
    \includegraphics[width=0.8\linewidth]{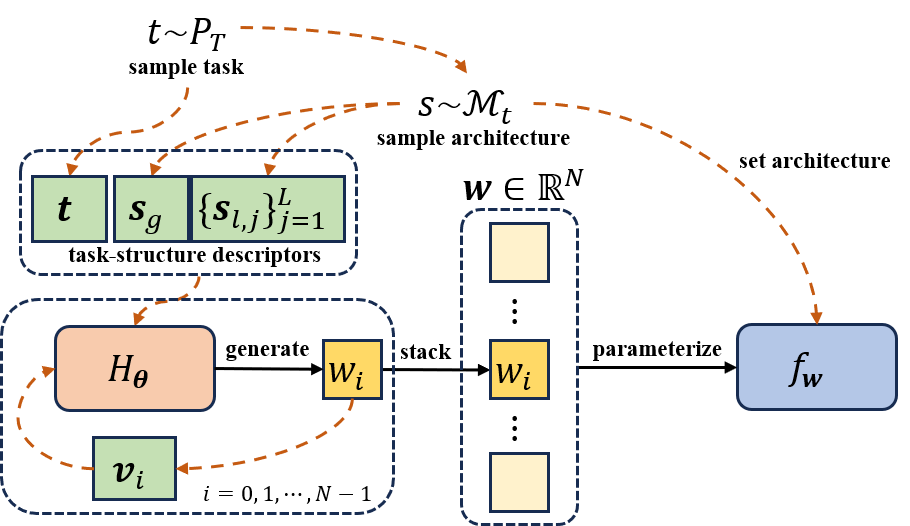}
    \caption{Method overview. We sample the task and architecture of the base model $f_{\mathbf{w}}$ and collect their task-structure descriptors. For each parameter in $f_{\mathbf{w}}$, we collect its index descriptor $\mathbf{v}_{i}$ and feed it into UHN $H_{\boldsymbol{\theta}}$ along with the task-structure descriptors to obtain each individual weight $w_{i}$ of $f$. These weights are then stacked into the weight vector $\mathbf{w}$, which parameterizes $f_{\mathbf{w}}$.}
    \label{fig:method-overview}
\end{figure}

Given a target specification (base model architecture and task), UHN generates each scalar parameter by querying a single fixed-architecture generator $H_{\boldsymbol{\theta}}$ with deterministic descriptors (similar in spirit to implicit neural representations). \cref{fig:method-overview} illustrates the weight generation process.

For a base model $f_{\mathbf{w}}$ with $N$ parameters, we adopt a hierarchical view of its architecture: a \emph{model} consists of $L$ \emph{layers}, and each layer consists of several \emph{components} (e.g., a linear layer has a weight matrix and a bias vector).

The key idea is to move all target-specificity into the \emph{inputs} to $H_{\boldsymbol{\theta}}$. Intuitively, to generate a scalar weight we need to tell the generator (i) \emph{which} parameter we are generating (its location/type within the architecture), and (ii) \emph{what} model and task that parameter belongs to (global and layer-level structure, plus task information). Concretely, for each parameter indexed by $i$, we form the following deterministic conditioning descriptors (detailed specifications in \cref{details-identifiers}) from the specified architecture and task description:
\begin{itemize}
    \item \textbf{Index descriptor} $\mathbf{v}_{i}$, encoding a parameter’s \emph{structural} location and type (e.g., layer ID, component type, intra-component coordinates).
    \item \textbf{Global structure descriptor} $\mathbf{s}_g$, encoding global architectural metadata (e.g., model family, depth, and other configuration-level attributes).
    \item \textbf{Local structure descriptors} $\{\mathbf{s}_{\ell,j}\}_{j=1}^{L}$, encoding layer-level metadata (e.g., layer type and input/output dimensions) for $L$ layers respectively.
    \item \textbf{Task descriptor} $\mathbf{t}$, encoding task-level information (e.g., task type and dataset).
\end{itemize}

UHN $H_{\boldsymbol{\theta}}$ predicts the value of each parameter based on the descriptors
\begin{equation}
w_{i}
= H_{\boldsymbol{\theta}}\big(\mathbf{v}_{i},\, \mathbf{s}_g,\, \{\mathbf{s}_{\ell,j}\}_{j=1}^{L},\, \mathbf{t}\big).
\label{eq:uhn-generate}
\end{equation}
Crucially, changing the target model or task changes only the \emph{queries} (the descriptors), not the architecture of $H_{\boldsymbol{\theta}}$.

\paragraph{Example.}
Consider a base model with a single linear layer (no bias) of shape $d_{\mathrm{in}}\times d_{\mathrm{out}}$ for MNIST \cite{lecun1998gradient} classification. For each scalar weight connecting input unit $p$ to output unit $q$, we construct $\mathbf{v}_{i}$ from coordinates such as $(p,q)$ (together with a layer/component code), set $\mathbf{s}_g$ to encode global configuration (e.g., model family and depth) and set the single-layer local code $\mathbf{s}_{\ell,1}$ to include $(d_{\mathrm{in}},d_{\mathrm{out}})$ and the layer type, and set $\mathbf{t}$ to encode the task/dataset. We query $H_{\boldsymbol{\theta}}$ for all parameters (efficiently via batching) and then reshape the outputs into $d_{\mathrm{in}}\times d_{\mathrm{out}}$ to parameterize the base model.

\subsubsection{Motivation}
Many existing hypernetworks are \emph{entangled} with the target model: their output interface (e.g., output dimensionality, the number of per-layer/per-chunk embeddings, or specialized component heads) is tied to the base architecture \cite{hedlin2025hypernet, von2019continual}.
Consequently, changing the target architecture often changes the hypernetwork parameterization and requires redesign \cite{chen2022transformers, navon2020learning}.
We also provide a scaling analysis of chunk-based hypernetworks \cite{von2019continual}, showing how the hypernetwork size can scale with the target model (\cref{app:chunked-hypernet-scaling}).
UHN instead fixes the generator parameterization $\boldsymbol{\theta}$ and moves target-specificity into deterministic conditioning inputs, enabling a single UHN to support single-model, multi-model, multi-task, and recursive settings without modifying the generator itself.

% Draw method overview

%%%%%%%%%%%%%%%%%%%%%%%%%%%%%%%%
% Universal Generation Capability
%%%%%%%%%%%%%%%%%%%%%%%%%%%%%%%%
\subsubsection{Universal generation capability}
We next describe how UHN covers several generation settings. The unifying theme is that we keep a single generator $H_{\boldsymbol{\theta}}$ fixed, and obtain different behaviors \emph{only} by changing the descriptors we query it with.

\paragraph{Single-model, single-task.}
Fix the target specification $(\mathbf{s}_g,\{\mathbf{s}_{\ell,j}\}_{j=1}^{L},\mathbf{t})$ and generate all parameters by evaluating $H_{\boldsymbol{\theta}}$ on the corresponding index descriptors.
In the simplest (non-conditional) variant, we can drop structure/task conditioning and use $w_{i}=H_{\boldsymbol{\theta}}(\mathbf{v}_{i})$.

\paragraph{Multi-model, single-task.}
Fix the task code $\mathbf{t}$, and vary the architecture descriptors $(\mathbf{s}_g,\{\mathbf{s}_{\ell,j}\}_{j=1}^{L})$ across different base-model architectures. For each architecture, we generate its parameters with the same $H_{\boldsymbol{\theta}}$.

\paragraph{Multi-task.}
Vary the task code $\mathbf{t}$ across tasks, and for each task vary the architecture descriptors $(\mathbf{s}_g,\{\mathbf{s}_{\ell,j}\}_{j=1}^{L})$ across the corresponding task-specific model architectures.
UHN then generates a task-specific model by conditioning on both the chosen task and architecture.

\paragraph{Recursive generation.}
Treat a hypernetwork itself as a target model.
Given the specification of a target hypernetwork $H_{k+1}$ and its index descriptors $\mathbf{v}_{k,i}$, we generate its parameters $\boldsymbol{\theta}_{k+1}$ using the same procedure as for base-model weights. The overall parameter generation process from the root UHN $H_0$ to the base model $f$ generated by the intermediate UHN $H_K$ can be denoted as a recursion chain $H_0\rightarrow\cdots\rightarrow H_k\rightarrow\cdots\rightarrow H_K\rightarrow f$.
This recursion highlights the central payoff of UHN: once the descriptor construction is defined for a model family, the same machinery can be applied repeatedly without introducing architecture-specific heads or changing output dimensionalities.

%%%%%%%%%%%%%%%%%%%%%%%%%%%%%%%%
% UHN Architecture
%%%%%%%%%%%%%%%%%%%%%%%%%%%%%%%%
\subsubsection{UHN architecture}
\label{sec:encoding-arch}
This section describes the UHN forward pass as a simple two-stage pipeline: (i) deterministically encode the raw descriptors into fixed-length features, and (ii) map these features to a scalar weight with a learnable generator network.
For inference pseudocode, see \cref{alg:uhn-infer}.

\paragraph{Stage 1: encode descriptors.}
Given a target specification (architecture/task), we deterministically construct descriptors $(\mathbf{v}_i,\mathbf{s}_g,\{\mathbf{s}_{\ell,j}\}_{j=1}^{L},\mathbf{t})$ for each parameter index~$i$.
At a high level, Stage~1 applies Gaussian Fourier features \cite{tancik2020fourier} to mitigate spectral bias (neural networks tend to fit low-frequency functions first), making it easier for the generator to represent higher-frequency structure in the weight field. Concretely, we form per-layer task-structure descriptors $\mathbf{u}_j=[\mathbf{s}_g;\mathbf{t};\mathbf{s}_{\ell,j}]$ and compute fixed (non-learned) encodings for $\mathbf{v}_i$ and $\mathbf{u}_j$ as follows.
\begin{equation}
\boldsymbol{\phi}_i \;=\; \gamma_{\mathbf{B}_{\mathbf{v}}}(\hat{\mathbf{v}}_i),
\qquad
\boldsymbol{\psi}_j \;=\; \gamma_{\mathbf{B}_{\mathbf{u}}}(\hat{\mathbf{u}}_j).
\end{equation}
For each descriptor vector $\mathbf{x}\in\{\mathbf{v}_i,\mathbf{u}_j\}$, we first compute its elementwise-normalized version $\hat{\mathbf{x}}$ (details in \cref{details-method-normal}) and then apply the Gaussian Fourier feature map
\begin{equation}
\gamma_{\mathbf{B}}(\hat{\mathbf{x}}) = \big[\cos(\mathbf{B}\hat{\mathbf{x}})^{\top},\; \sin(\mathbf{B}\hat{\mathbf{x}})^{\top}\big]^{\top},
\end{equation}
where $\hat{\mathbf{x}}\in\mathbb{R}^{n}$, $\mathbf{B}\in\mathbb{R}^{m\times n}$ has entries sampled i.i.d.\ from $\mathcal{N}(0,\sigma^2)$ once and then kept fixed, and hence $\gamma_{\mathbf{B}}(\hat{\mathbf{x}})\in\mathbb{R}^{2m}$.

\paragraph{Stage 2: map encodings to weights.}
The core of UHN $H_{\boldsymbol{\theta}}$ consists of (i) an \emph{index branch} that processes the per-parameter index feature $\boldsymbol{\phi}_i$, and (ii) an optional \emph{task-structure encoder} that aggregates the per-layer features $\{\boldsymbol{\psi}_j\}_{j=1}^{L}$ into a single global conditioning vector.

Concretely, the index branch is a multilayer perceptron (MLP) with pre-activation residual \cite{he2016identity} blocks: it maps $\boldsymbol{\phi}_i$ through an input linear layer (output width $d$) followed by 2 pre-activation residual blocks, where each block consists of two (ReLU \cite{nair2010rectified} $\rightarrow$ LayerNorm \cite{ba2016layer} $\rightarrow$ Linear) layers with hidden width $d$ (no bottleneck) and a shortcut connection from the block input to its output.

When conditioning on architecture/task, we compute a task-structure feature via the task-structure encoder: a single-layer Transformer encoder \cite{vaswani2017attention} ($h$ heads; feed-forward dimension equal to the encoder input dimension, i.e., $d_{\mathrm{ff}}=\dim(\boldsymbol{\psi}_j)$; no dropout) applied to $\{\boldsymbol{\psi}_j\}_{j=1}^{L}$, followed by mean pooling across the layer/token dimension and a 2-layer MLP (Linear $\rightarrow$ ReLU $\rightarrow$ Linear, with hidden and output width $d$; last linear layer zero-initialized).
We add this task-structure feature to the index-branch representation (after the residual blocks), and pass the fused representation through a ReLU and a final linear layer to predict the scalar weight $w_{i}$.
Additional architecture hyperparameters are given in \cref{details-method-sweep}.

%%%%%%%%%%%%%%%%%%%%%%%%%%%%%%%%
% Training Procedure
%%%%%%%%%%%%%%%%%%%%%%%%%%%%%%%%
\subsection{Training procedure}
\label{sec:training}
We train UHN using a unified procedure that covers the single-model, multi-model, multi-task, and recursive settings (see \cref{alg:uhn-train}).
At a high level, each iteration samples a target specification (architecture and task), uses the root UHN to generate the corresponding parameters, evaluates the task loss of the resulting base model, and backpropagates gradients through the entire differentiable generation path to update the root parameters.
In practice, we optionally begin with an initialization phase that matches simple parameter statistics (see \cref{alg:uhn-init}), followed by standard task training.
The following parts detail how we sample architectures/tasks during training, and how we perform initialization.

\subsubsection{Model architecture sampling}
\label{details-method-model-sample}
Let $M$ be a set of model architectures, and let $\mathcal{M}$ be the uniform distribution over $M$. At each training iteration, we sample an architecture $s\sim \mathcal{M}$ (with $s\in M$), which is represented by the structure descriptors $(\mathbf{s}_g,\{\mathbf{s}_{\ell,j}\}_{j=1}^{L})$.

\subsubsection{Task sampling}
\label{details-method-task-sample}
Let $\mathcal{T}$ denote a set of tasks (each task is defined by a task type and a dataset), and let $P_T$ denote a distribution over tasks.
At each training iteration, we first sample a task $t\sim P_T$ (with $t\in\mathcal{T}$), which is represented as task descriptor $\mathbf{t}$, and then sample an architecture $s\sim \mathcal{M}_{t}$ from the corresponding task-conditional model structure distribution.

\subsubsection{Initialization strategy}
Before the main training phase, we optionally stabilize optimization with an initialization phase that plays a similar role to standard weight initialization, by matching simple parameter statistics (e.g., mean/variance) to common initialization choices.
Let $G$ denote the set of all generated-model components, and let $g\in G$ denote a particular component (e.g., a weight matrix or bias vector) with empirical mean $\mu(g)$ and standard deviation $\sigma(g)$. We specify desired target statistics $(\mu^*(g),\sigma^*(g))$ for each component (see \cref{details-method-init-targets}) and minimize
\begin{equation}
\label{init_loss}
\mathcal{L}_{\mathrm{init}}=\frac{1}{2|G|}\sum_{g\in G}\Big[\Big(\mu(g)-\mu^*(g)\Big)^2+\Big(\sigma(g)-\sigma^*(g)\Big)^2\Big].
\end{equation}

\paragraph{Initialization for recursive generation.}
\label{details-method-recursive-init}
For recursive chains $H_0\rightarrow\cdots\rightarrow H_k\rightarrow\cdots\rightarrow H_K\rightarrow f$, we initialize the chain in a top-down fashion: we first initialize the parameters of $H_1$ (the model generated by $H_0$), then $H_2$, and so on, until finally initializing the leaf/base model $f$ generated by $H_K$.
We implement this by assigning a fixed budget of $S_{\mathrm{lvl}}$ initialization steps to each level and selecting an active level index $k$ as
\begin{equation}
k=\min\big(K,\lfloor \mathrm{step}/S_{\mathrm{lvl}}\rfloor\big)
\end{equation}
where $\mathrm{step}$ denotes the (0-indexed) initialization iteration counter.
We interpret $k$ as follows: if $k<K$, then we initialize the parameters of $H_{k+1}$ (i.e., the model generated by $H_k$); if $k=K$, we initialize the parameters of the leaf/base model $f$. Thus, as the step count increases, initialization proceeds from shallow (early) levels (e.g., $H_1$) toward deeper (later) levels (e.g., $H_K$ and finally $f$).

%%%%%%%%%%%%%%%%%%%%%%%%%%%%%%%%
% Experiments
%%%%%%%%%%%%%%%%%%%%%%%%%%%%%%%%
\section{Experiments}
%%%%%%%%%%%%%%%%%%%%%%%%%%%%%%%%
% Experiments Setup
%%%%%%%%%%%%%%%%%%%%%%%%%%%%%%%%
We organize multiple experiments to validate claims of this paper.
Single-model experiments establish \emph{universality} across heterogeneous task--model pairs and test \emph{scalability} as target size grows under a fixed UHN.
Multi-model experiments test \emph{family-level generalization} to held-out architectures within a model family.
Multi-task experiments evaluate a \emph{single shared UHN} across heterogeneous datasets and modalities.
Finally, recursive experiments probe \emph{multi-level generation} by chaining UHNs.
We conduct ablations of key UHN design choices, including index encoding, hypernetwork capacity, the task-structure encoder, and initialization, as specified in \cref{app:ablations}.
\subsection{Experimental Setup}
\paragraph{Hardware and framework.}
All experiments are implemented in PyTorch \cite{paszke2019pytorch} and run on a single NVIDIA RTX~4090 GPU.
We enable automatic mixed precision (AMP, FP16) to reduce memory usage and speed up training.
\paragraph{Optimization.}
When training UHN, we use AdamW \cite{loshchilov2017decoupled} with a cosine learning-rate schedule \cite{loshchilov2016sgdr} for both initialization phase and main training phase. In the main training phase, we use $5$ linear learning-rate warmup \cite{goyal2017accurate} epochs for single-model and multi-model experiments, and $1000$ warmup steps for multi-task/recursive experiments. 
For classification tasks (image/graph/text), we use mini-batches of size $256$.
For formula regression, we use full-batch training (batch size equals the training-set size).
We do not use weight decay in any experiment.
\paragraph{Reporting and splits.}
We report mean $\pm$ standard deviation over multiple random seeds, with the exact seed counts stated per experiment.
Dataset splits, base-model architectures and model-family sampling details are provided in \cref{details-exp-single-model,details-exp-multi-model} for reproducibility.

%%%%%%%%%%%%%%%%%%%%%%%%%%%%%%%%
% Single Model Experiment
%%%%%%%%%%%%%%%%%%%%%%%%%%%%%%%%
\subsection{Single-Model Experiments}
\label{single-model-exp}
We first study the single-model setting: in each experiment, we fix one task and one target base architecture, and train a UHN to generate the weights of that target model.
We evaluate (i) universality across heterogeneous task--model pairs, and (ii) scalability to larger target models under a fixed UHN.
Because the task and target architecture are fixed, we omit UHN's task-structure encoder in this case.
Detailed base-model architectures and additional implementation details are provided in \cref{details-exp-single-model}.

\subsubsection{Universality}

\paragraph{Setup.}
We consider four types of tasks:
(i) image classification on MNIST \cite{lecun1998gradient} and CIFAR-10 \cite{krizhevsky2009learning} via multilayer perceptron (MLP) or convolutional neural network (CNN),
(ii) graph node classification on Cora, CiteSeer, and PubMed \cite{sen2008collective} via graph convolutional network (GCN) \cite{kipf2016semi} or graph attention network (GAT) \cite{velivckovic2017graph},
(iii) text classification on AG News \cite{zhang2015character} and IMDB \cite{maas2011learning} via Transformer \cite{vaswani2017attention}, and
(iv) formula regression via Kolmogorov-Arnold Network (KAN) on its 15 special-functions benchmark (listed in \cref{tab:single-model-universality-formula}) \cite{liu2024kan}.

\paragraph{Results.}
\label{sec:results-single-task}
We compare UHN-generated weights against direct training across vision, graph, text, and formula tasks (\cref{tab:single-model-universality-result} and \cref{tab:single-model-universality-formula}).
We train direct-training baselines (Direct) with AdamW and a cosine learning-rate schedule with $5$ warmup epochs.
For each setting, we report mean and standard deviation of \emph{test} performance over multiple random seeds (3 seeds for image/text/formula; 10 seeds for graph).
For classification tasks, we report test accuracy (higher is better); for formula regression, we report test RMSE (lower is better).

\begin{table*}[t]
\centering
\caption{Single-model universality results for classification tasks (image/graph/text). \#Params denotes the number of trainable parameters; Acc. denotes test accuracy. Due to space constraints, formula-regression results are reported in \cref{tab:single-model-universality-formula}.}
\label{tab:single-model-universality-result}
\scriptsize
\begin{tabular}{lllr r ll}
\toprule
Task type & Dataset & Model & \#Params (Direct) & \#Params (UHN) & Acc. (Direct) & Acc. (UHN) \\
\midrule
Image & MNIST & MLP & 118{,}282 & 158{,}613 & $0.9837 \pm 0.0007$ & $0.9841 \pm 0.0006$ \\
Image & MNIST & CNN-8 & 74{,}762 & 158{,}613 & $0.9938 \pm 0.0004$ & $0.9944 \pm 0.0010$ \\
Image & CIFAR-10 & CNN-20 & 269{,}034 & 612{,}117 & $0.8999 \pm 0.0027$ & $0.8993 \pm 0.0016$ \\
\midrule
Graph & Cora & GCN & 92{,}231 & 612{,}117 & $0.8172 \pm 0.0039$ & $0.7950 \pm 0.0069$ \\
Graph & CiteSeer & GCN & 237{,}446 & 612{,}117 & $0.6996 \pm 0.0051$ & $0.6529 \pm 0.0190$ \\
Graph & PubMed & GCN & 32{,}259 & 612{,}117 & $0.7691 \pm 0.0028$ & $0.7815 \pm 0.0075$ \\
Graph & Cora & GAT & 92{,}373 & 612{,}117 & $0.8132 \pm 0.0094$ & $0.7981 \pm 0.0091$ \\
Graph & CiteSeer & GAT & 237{,}586 & 612{,}117 & $0.6999 \pm 0.0051$ & $0.6819 \pm 0.0068$ \\
Graph & PubMed & GAT & 33{,}779 & 612{,}117 & $0.7700 \pm 0.0033$ & $0.7801 \pm 0.0037$ \\
\midrule
Text & AG News & Transformer-2L & 378{,}372 & 612{,}117 & $0.9186 \pm 0.0008$ & $0.9099 \pm 0.0005$ \\
Text & IMDB & Transformer-1L & 377{,}858 & 612{,}117 & $0.8853 \pm 0.0007$ & $0.8638 \pm 0.0005$ \\
\bottomrule
\end{tabular}
\end{table*}

\paragraph{Analysis.}
The MNIST experiments serve as a proof-of-concept and use a smaller-capacity UHN.
All other experiments share the same UHN architecture and capacity, and the only experiment-specific component is the UHN input descriptors specified by target architectures and tasks. Across different settings, performance remains close to the direct-training baseline despite substantial variation in both modality (vision/graph/text/formula) and target model family.

Overall, these results indicate that a single UHN can parameterize heterogeneous model classes with competitive test performance, supporting the universality claim of the proposed formulation.

\subsubsection{Scalability}
\label{sec:scalability}

\paragraph{Setup.}
We test scalability: whether a fixed-capacity UHN can generate the weights of increasingly large target models without changing the hypernetwork.
We instantiate this study on CIFAR-10 by increasing the generated network depth (CNN-20 $\rightarrow$ CNN-32/44/56).

\paragraph{Results.}
We compare UHN against (i) Direct training (Direct), where each target model is optimized end-to-end, and (ii) two embedding-based hypernetwork baselines: the hypernetwork of Ha et al.~\cite{ha2016hypernetworks} (HA) and the chunked hypernetwork of Oswald et al.~\cite{von2019continual} (Chunked).
We train direct-training, HA and Chunked baselines with AdamW and a cosine learning-rate schedule with $5$ warmup epochs.
Following common practice for HA-style hypernetworks in CNNs, HA generates only the convolution kernels, while other parameters are directly optimized.
We choose these representative embedding-based hypernetworks to highlight a common scalability pattern: per-target embeddings cause the hypernetwork parameter count to grow as the target model size increases.
In contrast, UHN uses no per-target embeddings, so its parameter count is independent of the target model size.
\Cref{tab:single-model-scalability} reports CIFAR-10 test accuracy (mean $\pm$ std over 3 seeds) for the target CNN variants.

\begin{table*}[t]
    \centering
    \scriptsize
    \begin{tabular}{lrrrr rrrr}
        \toprule
        & \multicolumn{4}{c}{\#Params} & \multicolumn{4}{c}{Acc.} \\
        \cmidrule(lr){2-5} \cmidrule(lr){6-9}
        Model & Direct & HA & Chunked & UHN & Direct & HA & Chunked & UHN \\
        \midrule
        CNN-20 & 269{,}034 & 619{,}306 & 625{,}480 & 612{,}117 & $0.8999 \pm 0.0027$ & $0.9070 \pm 0.0039$ & $0.9010 \pm 0.0008$ & $0.8993 \pm 0.0016$ \\
        CNN-32 & 463{,}018 & 635{,}546 & 627{,}592 & 612{,}117 & $0.9069 \pm 0.0028$ & $0.9124 \pm 0.0021$ & $0.9010 \pm 0.0002$ & $0.9035 \pm 0.0026$ \\
        CNN-44 & 657{,}002 & 651{,}786 & 629{,}640 & 612{,}117 & $0.9076 \pm 0.0010$ & $0.9161 \pm 0.0017$ & $0.9063 \pm 0.0030$ & $0.9043 \pm 0.0031$ \\
        CNN-56 & 850{,}986 & 668{,}026 & 631{,}688 & 612{,}117 & $0.9043 \pm 0.0036$ & $0.9159 \pm 0.0004$ & $0.9027 \pm 0.0021$ & $0.9069 \pm 0.0015$ \\
        \bottomrule
    \end{tabular}
    \caption{Single-model scalability results on CIFAR-10. \#Params denotes the number of trainable parameters; Acc. denotes test accuracy.}
    \label{tab:single-model-scalability}
\end{table*}

\paragraph{Analysis}
UHN scales to larger CNN variants without changing the hypernetwork architecture: its parameter count stays fixed, while HA and Chunked grow with target size (via more embeddings).
In accuracy, UHN is comparable to direct training and Chunked, while HA performs best in this setting, potentially due to its more constrained parameterization (a linear mapping from embeddings to weights).

%%%%%%%%%%%%%%%%%%%%%%%%%%%%%%%%
% Multi Model Experiment
%%%%%%%%%%%%%%%%%%%%%%%%%%%%%%%%
\subsection{Multi-Model Experiments}
\label{sec:multi-model-experiment}
\subsubsection{Setup}
We evaluate \emph{family-level} learning by training a single UHN to generate many architectures from a predefined model family, and testing whether it generalizes to held-out architectures from the same family.
We consider three CNN families on CIFAR-10 (CNN Mixed Depth, CNN Mixed Width, and CNN Mixed Depth\,$\times$\,Width) and one Transformer family on AG News (Transformer Mixed).
Here, ``Mixed'' indicates that we vary architectural hyperparameters within the family (CNN depth/width; Transformer depth/width-related settings, etc.); full model family details are provided in \cref{model-sampling-details}.
For each experiment, we sample a finite model set $M$ once from its model family and split it into $M_{\mathrm{train}}$ and $M_{\mathrm{test}}$ for training and held-out evaluation.
We additionally use a subset $M'_{\mathrm{train}} \subset M_{\mathrm{train}}$ for hold-in evaluation, and reserve another exclusive subset $M_{\mathrm{val}}$ for validation during hyperparameter selection.
We use the task-structure encoder in UHN to encode the base-model structure.
Model set splitting and hyperparameter selection details are provided in \cref{details-exp-multi-model}.

\subsubsection{Results}
To measure generalization within a family, we report test accuracy averaged over architectures in $M'_{\mathrm{train}}$ (``seen'') and $M_{\mathrm{test}}$ (``unseen'').
\cref{tab:multi-model-result} shows the mean and standard deviation of these averages over $n=3$ random seeds (each seed trains a separate UHN instance).

\begin{table*}[t]
    \centering
    \scriptsize
    \begin{tabular}{lrrrrrr}
        \toprule
        & \multicolumn{4}{c}{Model statistics} & \multicolumn{2}{c}{Acc.} \\
        \cmidrule(lr){2-5} \cmidrule(lr){6-7}
        Model family & \#Models & Avg. \#Params & Max \#Params & UHN \#Params & Seen & Unseen \\
        \midrule
        CNN Mixed Depth & 100 & 366{,}995 & 463{,}018 & 663{,}151 & $0.8842 \pm 0.0031$ & $0.8430 \pm 0.0023$ \\
        CNN Mixed Width & 500 & 624{,}987 & 1{,}072{,}586 & 663{,}151 & $0.9145 \pm 0.0014$ & $0.9145 \pm 0.0013$ \\
        CNN Mixed Depth $\times$ Width & 1000 & 734{,}206 & 1{,}372{,}110 & 663{,}151 & $0.9038 \pm 0.0012$ & $0.9040 \pm 0.0016$ \\
        Transformer Mixed & 1000 & 523{,}650 & 1{,}086{,}212 & 663{,}151 & $0.9063 \pm 0.0004$ & $0.9066 \pm 0.0002$ \\
        \bottomrule
    \end{tabular}
    \caption{Multi-model results. \#Models is the size of the model set $M$. Avg./Max \#Params are the average/maximum number of trainable parameters over base models in $M$. UHN \#Params is the number of trainable parameters of UHN. ``Seen'' averages test accuracy over $M'_{\mathrm{train}}$ and ``Unseen'' averages test accuracy over $M_{\mathrm{test}}$.}
    \label{tab:multi-model-result}
\end{table*}

\subsubsection{Analysis}
These results indicate that a single UHN can jointly parameterize a diverse set of architectures within each family while maintaining competitive performance (relative to the single-model setting in \cref{single-model-exp}).
Across all four experiments, we reuse the same UHN architecture even as the target models vary substantially in size (up to $\sim 1.37\times 10^6$ parameters for CNNs and $\sim 1.09\times 10^6$ parameters for Transformers).
Seen and unseen accuracies are close for CNN Mixed Width, CNN Mixed Depth\,$\times$\,Width, and Transformer Mixed, suggesting good generalization to held-out architectures.
The scatter plots in \cref{fig:model-family-scatter} further support that UHN can generate a family of architectures with diverse sizes and performance while preserving similar seen/unseen trends.

CNN Mixed Depth exhibits a larger gap between seen and unseen performance.
From the scatter plot (\cref{fig:model-family-scatter}), this gap is largely driven by a single unseen outlier (a model that is deeper than all seen models) with severely degraded accuracy; because $|M_{\mathrm{test}}|=20$, this single failure has a noticeable impact on the test accuracy averaged over $M_{\mathrm{test}}$.
One plausible explanation is that the smaller training model set ($|M_{\mathrm{train}}|=80$) under-represents rare ``edge'' configurations, making extrapolation to extreme depths more difficult.

%%%%%%%%%%%%%%%%%%%%%%%%%%%%%%%%
% Multi Task Experiment
%%%%%%%%%%%%%%%%%%%%%%%%%%%%%%%%
\subsection{Multi-Task Experiment}
\label{details-exp-multi-task}
\subsubsection{Setup}
We test the central multi-task claim of UHN: a single fixed-capacity hypernetwork can be trained once and then used to generate multiple task-specific base models across heterogeneous datasets and modalities.
We consider six task--model pairs spanning vision, graphs, text, and scientific regression: MLP on MNIST, CNN-44 on CIFAR-10, GCN on Cora, GAT on PubMed, Transformer-2L on AG News, and KAN-g5 on kv, one of the 15 special functions in the KAN benchmark.
Since the multi-model setting already evaluates varying base architectures within a single task, here we fix one base architecture per task to isolate the effect of heterogeneous multi-task training.
We use the task-structure encoder in UHN to encode the base-model structure and the task identity.

At each training step, we sample a task $t\sim P_T$ and train on that task; specifically, $P_T$ assigns probabilities: CIFAR-10 (0.55), AG News (0.18), kv (0.11), MNIST (0.08), Cora (0.04), and PubMed (0.04).
Details of hyperparameter selection are provided in \cref{app:details-exp-multi-task}.

\subsubsection{Results}
After joint training, we use UHN to generate each task's base model and evaluate test performance on that task.
\cref{tab:multi-task-result} reports the mean and standard deviation over $n=3$ random seeds.
For classification tasks we report \emph{test accuracy} (higher is better); for formula regression we report \emph{test RMSE} (lower is better).
For reference, we also report (i) direct training (training each model end-to-end) and (ii) a single-task UHN trained separately per task; both baselines are taken from the corresponding single-model experiments (\cref{sec:results-single-task}).

\begin{table*}[t]
    \centering
    \scriptsize
    \begin{tabular}{lll lll}
        \toprule
        Task type & Dataset & Model & Perf. (Direct) & Perf. (UHN Single-task) & Perf. (UHN Multi-task) \\
        \midrule
        Image & MNIST & MLP & $0.9837 \pm 0.0007$ & $0.9841 \pm 0.0006$ & $0.9786 \pm 0.0012$ \\
        Image & CIFAR-10 & CNN-44 & $0.9076 \pm 0.0010$ & $0.9043 \pm 0.0031$ & $0.8927 \pm 0.0011$ \\
        Graph & Cora & GCN & $0.8172 \pm 0.0039$ & $0.7950 \pm 0.0069$ & $0.7930 \pm 0.0053$ \\
        Graph & PubMed & GAT & $0.7700 \pm 0.0033$ & $0.7801 \pm 0.0037$ & $0.7697 \pm 0.0031$ \\
        Text & AG News & Transformer-2L & $0.9186 \pm 0.0008$ & $0.9099 \pm 0.0005$ & $0.9062 \pm 0.0005$ \\
        \midrule
        Formula (RMSE) & kv & KAN-g5 & $0.0211 \pm 0.0126$ & $0.0104 \pm 0.0048$ & $0.0172 \pm 0.0112$ \\
        \bottomrule
    \end{tabular}
    \caption{Multi-task results. Perf. denotes test accuracy for classification (higher is better) and test RMSE for formula regression (lower is better). ``UHN Single-task'' denotes a separate UHN trained per task; ``UHN Multi-task'' denotes one shared UHN trained jointly on all tasks. Direct and UHN Single-task baselines are taken from \cref{sec:results-single-task}.}
    \label{tab:multi-task-result}
\end{table*}

\subsubsection{Analysis}
Overall, a single multi-task UHN generates reasonable task-specific models across all six datasets, with performance close to direct training and to a dedicated single-task UHN.
Across classification tasks, the multi-task UHN remains close to both direct training and the single-task UHN; the largest gap appears on CIFAR-10 ($0.9076$ direct / $0.9043$ single-task UHN vs. $0.8927$ multi-task UHN).
On graph tasks, the multi-task UHN matches the single-task UHN on Cora within error bars and is slightly lower on PubMed.
For formula regression, multi-task training increases RMSE relative to the single-task UHN (from $0.0104$ to $0.0172$), suggesting strong cross-task interference for this objective.

We attribute the remaining gap to cross-task gradient interference and to uneven task exposure during training (since the expected number of updates per task scales with $P_T$).
We expect that gradient conflict mitigation (e.g., gradient surgery \cite{yu2020gradient}) and task-adaptive sampling strategy could further reduce interference, which we leave for future work.
These results support the claim that UHN can serve as a single shared generator in heterogeneous multi-task settings without changing the hypernetwork architecture.

%%%%%%%%%%%%%%%%%%%%%%%%%%%%%%%%
% Recursive Experiment
%%%%%%%%%%%%%%%%%%%%%%%%%%%%%%%%
\subsection{Recursive Experiments}
\label{details-exp-recursive-task}

\begin{table}[t]
    \centering
    \scriptsize
    \begin{tabular}{ll}
        \toprule
        Setting & Acc. \\
        \midrule
        Direct training & $0.9837 \pm 0.0007$ \\
        UHN (no recursion) & $0.9841 \pm 0.0006$ \\
        \midrule
        UHN rec. ($K=1$) & $0.9825 \pm 0.0007$ \\
        UHN rec. ($K=2$) & $0.9795 \pm 0.0011$ \\
        UHN rec. ($K=3$) & $0.9741 \pm 0.0021$ \\
        \bottomrule
    \end{tabular}
    \caption{Recursive-task results on MNIST. ``Direct training'' and ``UHN (no recursion)'' are taken from the single-model MNIST MLP experiment (\cref{sec:results-single-task}). ``UHN rec.'' reports the accuracy of the leaf MLP generated via a recursion chain of depth $K$. Acc. denotes test accuracy.}
    \label{tab:recursive-task-perf}
\end{table}

\subsubsection{Setup}
We study \emph{recursive weight generation} to test whether UHN can stably generate another UHN whose outputs are then used to instantiate a downstream task model.
This experiment is complementary to the multi-model and multi-task settings and isolates the behavior of multi-level generation.

We consider recursion chains of the form
\begin{equation}
H_0\rightarrow\cdots\rightarrow H_k\rightarrow\cdots\rightarrow H_K\rightarrow f,
\end{equation}
where each arrow denotes ``generate the weights of.'' Here all UHNs $H_k$ in the chain use the task-structure encoder, and $H_k$ for $k\ge 1$ are \emph{generated} UHNs. Concretely, $H_0$ (the root UHN) generates the weights of $H_1$; $H_1$ generates $H_2$; and so on until $H_K$ generates the leaf base model $f$ (an MLP for MNIST classification).
We evaluate recursion depths up to $K=3$.
We use standard stabilization techniques for recursive training (e.g., gradient clipping \cite{pascanu2013difficulty}); the generated-UHN architecture and additional implementation details are provided in \cref{app:recursive-details}.
Our recursive experiments are designed to establish feasibility and stability up to three generated UHN levels. We do not claim monotonic depth-scaling performance, because settings deeper than $K=3$ were not exhaustively tuned and likely require depth-specific optimization choices.

\subsubsection{Results}
We report MNIST \emph{test accuracy} for the leaf task model generated at the end of the recursion chain (\Cref{tab:recursive-task-perf}).
Results are averaged over 3 random seeds and reported as mean $\pm$ standard deviation.

\subsubsection{Analysis}
Recursive generation remains numerically stable up to depth $K=3$ under our training protocol.
Performance degrades gradually as recursion depth increases, which is expected since deeper chains compound approximation error: each generated hypernetwork must (i) be representable by its parent and (ii) be optimized through a longer differentiable generation path.
Nevertheless, even at $K=3$ the resulting MNIST accuracy remains close to the direct-training baseline, supporting the feasibility of multi-level recursive weight generation.

%%%%%%%%%%%%%%%%%%%%%%%%%%%%%%%%
% Discussion
%%%%%%%%%%%%%%%%%%%%%%%%%%%%%%%%
\section{Discussion}
\label{discussion}
\paragraph{From architectural coupling to a universal generator.}
The central implication of our results is that hypernetwork design can be decoupled from base-model design.
By mapping fixed parameter descriptors (index, structure, and task signals) to weights through one shared function, UHN avoids the architecture-specific output parameterization that often forces redesign when the target model changes \cite{chen2022transformers, navon2020learning}.
This makes a single trained UHN usable across heterogeneous model families and task domains, including settings where each task uses a different backbone.

\paragraph{Scalability and practical transfer.}
Because UHN uses deterministic descriptors rather than learned per-layer/per-block embedding sets \cite{hedlin2025hypernet, von2019continual}, scaling to diverse architectures does not require changing the generator architecture itself.
Our multi-model and multi-task results indicate that this shared parameter generator captures reusable structure across models while remaining competitive with direct training baselines on diverse benchmarks.
The recursive experiments further suggest that this formulation is compositional: one UHN can generate another, with stable generation up to three intermediate UHNs under our training protocol.

Extended discussion of shared structure across tasks/models, initialization effects, limitations, and future directions is provided in \cref{app:discussion}.

%%%%%%%%%%%%%%%%%%%%%%%%%%%%%%%%
% Conclusion
%%%%%%%%%%%%%%%%%%%%%%%%%%%%%%%%
\section{Conclusion}
\label{conclusion}
Universal Hypernetwork (UHN) shows that hypernetwork design can be made architecture-agnostic: a single generator can produce trainable parameters for heterogeneous target models across tasks.
By modeling weights as a shared function of parameter, architecture, and task descriptors, UHN removes the need to redesign the hypernetwork when the base model changes and enables unified multi-model and multi-task generation.
Across diverse benchmarks, UHN remains competitive while also supporting recursive generation, with stable chains of up to three intermediate UHNs.
Overall, these results position UHN as a practical step toward general-purpose, reusable hypernetworks.

\bibliography{references}
\bibliographystyle{icml2026}

%%%%%%%%%%%%%%%%%%%%%%%%%%%%%%%%%%%%%%%%%%%%%%%%%%%%%%%%%%%%%%%%%%%%%%%%%%%%%%%
%%%%%%%%%%%%%%%%%%%%%%%%%%%%%%%%%%%%%%%%%%%%%%%%%%%%%%%%%%%%%%%%%%%%%%%%%%%%%%%
% APPENDIX
%%%%%%%%%%%%%%%%%%%%%%%%%%%%%%%%%%%%%%%%%%%%%%%%%%%%%%%%%%%%%%%%%%%%%%%%%%%%%%%
%%%%%%%%%%%%%%%%%%%%%%%%%%%%%%%%%%%%%%%%%%%%%%%%%%%%%%%%%%%%%%%%%%%%%%%%%%%%%%%
\newpage
\appendix
\onecolumn

%%%%%%%%%%%%%%%%%%%%%%%%%%%%%%%%
% Appendix: Additional Related Work
%%%%%%%%%%%%%%%%%%%%%%%%%%%%%%%%
\section{Additional Related Work}
\label{app:additional-related-work}
This section complements \cref{related-work} by summarizing additional literature that provides context for UHN, with emphasis on multi-task hypernetworks \cite{navon2020learning} and implicit neural representations \cite{essakine2024we}.

\subsection{Multi-task hypernetworks}
In the multi-task setting, prior work has used hypernetworks to generate task-specific parameters in order to encourage parameter sharing and improve aggregate performance across tasks \cite{tay2021hypergrid, wang2025iop, zhao2020meta, mahabadi2021parameter, shamsian2021personalized, liu2022polyhistor, zhao2024breaking, mueller2024mothernet, sun2025hypertasr, nguyen2025framework}.
However, many approaches assume a shared base-network architecture across tasks and condition the hypernetwork on a task representation (e.g., a task embedding) to modulate generated weights \cite{von2019continual, navon2020learning}.
As a result, they do not directly address the practical setting where different tasks require heterogeneous architectures (e.g., CNNs for vision and Transformers for language), with substantially different inputs, outputs, and inductive biases \cite{beck2023hypernetworks, hoang2023improving}.
These limitations motivate UHN's task-and-architecture conditioning for heterogeneous multi-task generation.

\subsection{Implicit Neural Representations (INRs)}
Implicit neural representations (INRs) model signals as continuous coordinate-to-value functions parameterized by neural networks \cite{essakine2024we}.
Most INR research improves this mapping through alternative activations \cite{sitzmann2020implicit, saragadam2023wire} and more effective coordinate encodings \cite{tancik2020fourier, mildenhall2021nerf, muller2022instant}.

Applying INRs to neural-network weight generation can be viewed as a natural extension.
However, this direction has been less explored and has historically lagged direct training in performance \cite{ha2016hypernetworks, stanley2009hypercube}.
Related work also studies hypernetworks that generate INRs, rather than INRs that generate weights of conventional networks \cite{klocek2019hypernetwork, chen2022transformers, pilligua2025hypernvd}.

UHN adapts the INR perspective to weight generation by modeling network parameters as a function over parameter indices.
Concretely, UHN maps a deterministic index descriptor to a scalar weight, so the same hypernetwork can be applied to different base architectures by enumerating different indices without changing generator architecture.
The choice of index encoding is critical: with suitable encoding, INR-style hypernetworks can be competitive with direct training.
Finally, we adopt conditional INR formulations \cite{park2021hypernerf, martin2021nerf} by conditioning on architecture and task descriptors, enabling one generator to produce weights across heterogeneous model structures and tasks.

%%%%%%%%%%%%%%%%%%%%%%%%%%%%%%%%
% Appendix for Method
%%%%%%%%%%%%%%%%%%%%%%%%%%%%%%%%
\section{Method}
We begin with a scaling intuition for chunked hypernetworks \cite{von2019continual}, then specify descriptor layouts, present inference/training/initialization algorithms, and close with descriptor normalization, base-model and recursive-hypernetwork structure definitions, initialization target statistics, and hyperparameter-selection details.

\subsection{Scaling analysis of chunked hypernetworks}
\label{app:chunked-hypernet-scaling}

\paragraph{Why learned chunk embeddings induce growth with target size.}
Many hypernetworks generate the target parameters in \emph{chunks} (e.g., per-layer, per-module, or fixed-size blocks), where each chunk is identified by a learned code/embedding and then mapped to the corresponding weights \cite{ha2016hypernetworks, hedlin2025hypernet}.
We analyze a generic chunked hypernetwork template \cite{von2019continual}.

Concretely, consider a chunked hypernetwork that partitions $N$ target parameters into $N_{\mathrm{chunks}}$ chunks of size $c$ (so $N=cN_{\mathrm{chunks}}$) and assigns each chunk $r$ a learned embedding $\mathbf{e}_r\in\mathbb{R}^{d_{\mathrm{emb}}}$.
Let $o$ be the hidden width of the hypernetwork core just before the final linear output layer, and let $N_{H,0}$ denote the number of hypernetwork parameters that do not scale with $(N,c,N_{\mathrm{chunks}})$.
Then the total number of hypernetwork parameters
\begin{equation}
N_H = N_{H,0} + o c + N_{\mathrm{chunks}} d_{\mathrm{emb}}.
\end{equation}
Since $N_{\mathrm{chunks}}=N/c$, the arithmetic--geometric mean inequality yields the lower bound
\begin{equation}
    o c + N_{\mathrm{chunks}} d_{\mathrm{emb}}
    = o c + \frac{N}{c}d_{\mathrm{emb}}
    \ge 2\sqrt{o d_{\mathrm{emb}}}\,\sqrt{N},
\end{equation}
so for fixed $(o,d_{\mathrm{emb}})$, such hypernetworks satisfy $N_H=\Omega(\sqrt{N})$.

The lower bound shows why many chunked hypernetworks grow with the target size.
On the other hand, our UHN corresponds to the per-weight regime ($c=1$ and $N_{\mathrm{chunks}}=N$) but replaces learned embeddings with deterministic descriptors (i.e., $d_{\mathrm{emb}}=0$), yielding $N_H = N_{H,0} + o$, which is independent of the base-model parameter count $N$.

\subsection{Descriptor specifications}
\label{details-identifiers}
This section documents the descriptor layouts used in our implementation.
Categorical attributes are deterministically mapped to integer IDs using a fixed implementation-defined mapping.
Inapplicable \emph{index} fields are set to $-1$, while inapplicable \emph{structure} fields are set to $0$.
In our experiments, the \emph{task} fields are always applicable. Detailed per-layer/per-model active-field conventions are provided in \cref{details-method-base-structure}.

\paragraph{Index descriptor $\mathbf{v}_i\in\mathbb{R}^{10}$ (\cref{tab:idx-layout}).}
The index descriptor specifies the location and type of an individual parameter.
\begin{table}[t]
    \centering
    \scriptsize
    \begin{tabular}{clp{0.60\linewidth}}
        \toprule
        Idx & Field & Description \\
        \midrule
        0 & layer\_idx & Layer index. \\
        1 & layer\_type & Layer type (categorical). \\
        2 & param\_type & Parameter type within the layer (categorical). \\
        3 & out\_idx & Output index / output channel / output embedding channel. \\
        4 & in\_idx & Input index / input channel / input embedding channel. \\
        5 & kernel\_h\_idx & Convolution kernel height index ($h$ coordinate). \\
        6 & kernel\_w\_idx & Convolution kernel width index ($w$ coordinate). \\
        7 & embedding\_idx & Token index. \\
        8 & sequence\_idx & Token position. \\
        9 & grid\_idx & KAN \cite{liu2024kan} grid/knots index. \\
        \bottomrule
    \end{tabular}
    \caption{Index descriptor layout $\mathbf{v}_i\in\mathbb{R}^{10}$.}
    \label{tab:idx-layout}
\end{table}

\paragraph{Global structure descriptor $\mathbf{s}_g\in\mathbb{R}^{6}$ (\cref{tab:sg-layout}).}
The global structure descriptor encodes architecture-level metadata.
\begin{table}[t]
    \centering
    \scriptsize
    \begin{tabular}{clp{0.60\linewidth}}
        \toprule
        Idx & Field & Description \\
        \midrule
        0 & model\_type & Model family type (categorical). \\
        1 & num\_layers & Number of layers. \\
        2 & cnn\_stage\_num & Number of stages (CNN). \\
        3 & num\_encoders & Number of encoder blocks (Transformer \cite{vaswani2017attention}). \\
        4 & num\_structure\_freqs & Number of frequencies for structure/task Fourier features \cite{tancik2020fourier} (Recursive UHN). \\
        5 & num\_index\_freqs & Number of frequencies for index Fourier features (Recursive UHN). \\
        \bottomrule
    \end{tabular}
    \caption{Global structure descriptor layout $\mathbf{s}_g\in\mathbb{R}^{6}$.}
    \label{tab:sg-layout}
\end{table}

\paragraph{Local structure descriptor $\mathbf{s}_{\ell,j}\in\mathbb{R}^{21}$ (\cref{tab:sl-layout}).}
The local descriptor describes layer $j$.
\begin{table}[t]
    \centering
    \scriptsize
    \begin{tabular}{clp{0.60\linewidth}}
        \toprule
        Idx & Field & Description \\
        \midrule
        0 & layer\_idx & Layer index. \\
        1 & layer\_type & Layer type (categorical). \\
        2 & bias\_type & Bias option (categorical). \\
        3 & norm\_type & Normalization option (categorical). \\
        4 & shortcut\_type & Shortcut \cite{he2016identity} option (categorical). \\
        5 & output\_size & Output dimension / channels / embedding dim. \\
        6 & input\_size & Input dimension / channels / embedding dim. \\
        7 & activation\_type & Activation function (categorical). \\
        8 & activation\_param & Activation parameter (e.g., leaky rate). \\
        9 & dropout\_rate & Dropout \cite{srivastava2014dropout} probability. \\
        10 & input\_pooling\_reshape\_type & Input pooling/reshape option (Linear layer; categorical). \\
        11 & group\_num & GroupNorm \cite{wu2018group} groups (Convolution layer). \\
        12 & kernel\_size & Kernel size (Convolution layer). \\
        13 & stage\_wise\_pooling\_type & Stage-wise pooling option (Convolution layer; categorical). \\
        14 & num\_heads & Number of heads (GAT \cite{velivckovic2017graph} / Multi-Head Attention (MHA) \cite{vaswani2017attention}). \\
        15 & head\_concat\_type & Head aggregation option (GAT; categorical). \\
        16 & embedding\_num & Vocabulary/embedding table size. \\
        17 & max\_sequence\_length & Maximum sequence length. \\
        18 & grid\_size & KAN grid size. \\
        19 & spline\_order & KAN spline order. \\
        20 & initialization\_type & Initialization option (categorical). \\
        \bottomrule
    \end{tabular}
    \caption{Local structure descriptor layout $\mathbf{s}_{\ell,j}\in\mathbb{R}^{21}$.}
    \label{tab:sl-layout}
\end{table}

\paragraph{Task descriptor $\mathbf{t}\in\mathbb{R}^{2}$ (\cref{tab:t-layout}).}
The task descriptor encodes task-level information (task type and dataset).
\begin{table}[t]
    \centering
    \scriptsize
    \begin{tabular}{clp{0.60\linewidth}}
        \toprule
        Idx & Field & Description \\
        \midrule
        0 & task\_type & Task type (categorical). \\
        1 & dataset\_type & Dataset type (categorical). \\
        \bottomrule
    \end{tabular}
    \caption{Task descriptor layout $\mathbf{t}\in\mathbb{R}^{2}$.}
    \label{tab:t-layout}
\end{table}

\noindent Putting the above together, a single generated parameter $w_{i}$ is determined by its index descriptor $\mathbf{v}_i$ (what parameter this is), the global structure descriptor $\mathbf{s}_g$ and a group of local structure descriptors $\{\mathbf{s}_{\ell,j}\}_{j=1}^{L}$ for a model with $L$ layers (what model it belongs to), and the task descriptor $\mathbf{t}$ (what task it serves),

\subsection{Algorithms}
\label{sec:appendix-algorithms}

We summarize UHN inference, training, and initialization in pseudocode (\cref{alg:uhn-infer,alg:uhn-train,alg:uhn-init}).

\begin{algorithm}[t]
\caption{Inference (with recursive generation)}
\label{alg:uhn-infer}
\begin{algorithmic}[1]
\STATE \textbf{Input:} root UHN parameters $\boldsymbol{\theta}$, recursion chain $H_0\rightarrow\cdots\rightarrow H_k\rightarrow\cdots\rightarrow H_K\rightarrow f$ 
\STATE Set $\boldsymbol{\theta}_{0} \leftarrow \boldsymbol{\theta}$ \COMMENT{parameters of $H_0$}
\FOR{$k=0$ \textbf{to} $K$}
    \STATE Gather the structure and task descriptors $(\mathbf{s}_{g,k},\{\mathbf{s}_{\ell,k,j}\}_{j=1}^{L_k},\mathbf{t}_k)$, and all index descriptors $\{\mathbf{v}_{k,i}\}_{i=1}^{N_k}$ for the target model $H_{k+1}$ when $k<K$ or $f$ when $k=K$
    \STATE Generate target model parameters $w_{k,i} \leftarrow H_{\boldsymbol{\theta}_{k}}\big(\mathbf{v}_{k,i},\mathbf{s}_{g,k},\{\mathbf{s}_{\ell,k,j}\}_{j=1}^{L_k},\mathbf{t}_k\big)$ for all $i\in\{1,\dots,N_k\}$
    \IF{$k<K$}
        \STATE Set $\boldsymbol{\theta}_{k+1} \leftarrow \mathbf{w}_{k}$ \COMMENT{parameters of generated $H_{k+1}$}
    \ELSE
        \STATE \textbf{return} $\mathbf{w}_{k}$ \COMMENT{parameters of generated $f$}
    \ENDIF
\ENDFOR
\end{algorithmic}
\end{algorithm}

\begin{algorithm}[t]
\caption{Training (with recursive generation)}
\label{alg:uhn-train}
\begin{algorithmic}[1]
\STATE \textbf{Input:} root UHN parameters $\boldsymbol{\theta}$, total training steps $S_{\mathrm{train}}$, initial task distribution $P_{T_0}$, recursive task distribution $P_{T_{k+1}\mid t_{k}}$, task-conditional model structure distribution $\mathcal{M}_{t_k}$
\STATE Initialize optimizer and learning-rate scheduler
\FOR{$\text{step}=0$ \textbf{to} $S_{\mathrm{train}}-1$}
    \STATE Sample root task $t_0\sim P_{T_0}$ and root structure $s_0\sim \mathcal{M}_{t_0}$
    \STATE Set $k\leftarrow 0$
    \WHILE{$t_k$ is recursive}
        \STATE Sample next task $t_{k+1}\sim P_{T_{k+1}\mid t_k}$ and structure $s_{k+1}\sim \mathcal{M}_{t_{k+1}}$
        \STATE $k\leftarrow k+1$
    \ENDWHILE
    \STATE Let $K\leftarrow k$; the sampled chain $(t_0,s_0),\ldots,(t_K,s_K)$ defines $H_0\rightarrow\cdots\rightarrow H_K\rightarrow f$
    \STATE Set $\boldsymbol{\theta}_{0} \leftarrow \boldsymbol{\theta}$ \COMMENT{parameters of $H_0$}
    \FOR{$k=0$ \textbf{to} $K$}
        \STATE Gather the structure and task descriptors $(\mathbf{s}_{g,k},\{\mathbf{s}_{\ell,k,j}\}_{j=1}^{L_k},\mathbf{t}_k)$, and all index descriptors $\{\mathbf{v}_{k,i}\}_{i=1}^{N_k}$ for the target model $H_{k+1}$ when $k<K$ or $f$ when $k=K$
        \STATE Generate target model parameters $w_{k,i} \leftarrow H_{\boldsymbol{\theta}_{k}}\big(\mathbf{v}_{k,i},\mathbf{s}_{g,k},\{\mathbf{s}_{\ell,k,j}\}_{j=1}^{L_k},\mathbf{t}_k\big)$ for all $i\in\{1,\dots,N_k\}$
        \IF{$k<K$}
            \STATE Set $\boldsymbol{\theta}_{k+1} \leftarrow \mathbf{w}_{k}$ \COMMENT{parameters of generated $H_{k+1}$}
        \ELSE
            \STATE Acquire the corresponding dataset $\mathcal{D}_{t_K}$ and loss function $\ell_{t_K}$ for leaf task $t_K$
            \STATE Sample a mini-batch $(\mathbf{X},\mathbf{Y})\sim \mathcal{D}_{t_K}$
            \STATE Compute loss $\mathcal{L} \leftarrow \ell_{t_K}\big(f_{\mathbf{w}_{k}}(\mathbf{X}),\mathbf{Y}\big)$
            \STATE Backpropagate $\mathcal{L}$ through the generation path and update $\boldsymbol{\theta}$
        \ENDIF
    \ENDFOR
\ENDFOR
\end{algorithmic}
\end{algorithm}

\begin{algorithm}[t]
\caption{Initialization (with recursive generation)}
\label{alg:uhn-init}
\begin{algorithmic}[1]
\STATE \textbf{Input:} root UHN parameters $\boldsymbol{\theta}$, total initialization steps $S_{\mathrm{init}}$, per-initialization-level steps $S_{\mathrm{lvl}}$, initial task distribution $P_{T_0}$, recursive task distribution $P_{T_{k+1}\mid t_{k}}$, task-conditional model structure distribution $\mathcal{M}_{t_k}$
\FOR{$\text{step}=0$ \textbf{to} $S_{\mathrm{init}}-1$}
    \IF{$\text{step} \bmod S_{\mathrm{lvl}} = 0$}
        \STATE Initialize (or reset) optimizer and learning-rate scheduler for the new initialization level
    \ENDIF
    \STATE Sample root task $t_0\sim P_{T_0}$ and root structure $s_0\sim \mathcal{M}_{t_0}$
    \STATE Set $k\leftarrow 0$
    \WHILE{$t_k$ is recursive}
        \STATE Sample next task $t_{k+1}\sim P_{T_{k+1}\mid t_k}$ and structure $s_{k+1}\sim \mathcal{M}_{t_{k+1}}$
        \STATE $k\leftarrow k+1$
    \ENDWHILE
    \STATE Let $K\leftarrow k$; the sampled chain $(t_0,s_0),\ldots,(t_K,s_K)$ defines $H_0\rightarrow\cdots\rightarrow H_K\rightarrow f$
    \STATE Set $\boldsymbol{\theta}_{0} \leftarrow \boldsymbol{\theta}$ \COMMENT{parameters of $H_0$}
    \STATE Set $r \leftarrow \min(K,\lfloor \text{step}/S_{\mathrm{lvl}} \rfloor)$ \COMMENT{current initialization level}    
    \FOR{$k=0$ \textbf{to} $r$}
        \STATE Gather the structure and task descriptors $(\mathbf{s}_{g,k},\{\mathbf{s}_{\ell,k,j}\}_{j=1}^{L_k},\mathbf{t}_k)$, and all index descriptors $\{\mathbf{v}_{k,i}\}_{i=1}^{N_k}$ for the target model $H_{k+1}$ when $k<K$ or $f$ when $k=K$
        \STATE Generate target model parameters $w_{k,i} \leftarrow H_{\boldsymbol{\theta}_{k}}\big(\mathbf{v}_{k,i},\mathbf{s}_{g,k},\{\mathbf{s}_{\ell,k,j}\}_{j=1}^{L_k},\mathbf{t}_k\big)$ for all $i\in\{1,\dots,N_k\}$
        \IF{$k<r$}
            \STATE Set $\boldsymbol{\theta}_{k+1} \leftarrow \mathbf{w}_{k}$ \COMMENT{parameters of generated $H_{k+1}$}
        \ELSE
            \STATE Compute initialization loss $\mathcal{L}_{\mathrm{init}}$ on $\mathbf{w}_{k}$ via \cref{init_loss}
            \STATE Backpropagate $\mathcal{L}_{\mathrm{init}}$ through the generation path and update $\boldsymbol{\theta}$
        \ENDIF
    \ENDFOR
\ENDFOR
\end{algorithmic}
\end{algorithm}

\subsection{Descriptor normalization}
\label{details-method-normal}
Before applying Gaussian Fourier features \cite{tancik2020fourier}, we normalize each scalar attribute of the descriptor vectors independently (in particular, of the index descriptor $\mathbf{v}_i$ and the task-structure descriptors $\mathbf{u}_j$). For a given scalar attribute $x\in[x_{\min},x_{\max}]$, where $x_{\min}$ and $x_{\max}$ are the minimum and maximum values estimated from the training distribution, we apply
\begin{equation}
\hat{x} = \frac{x-\mu}{\sigma},\qquad \mu = \frac{x_{\min}+x_{\max}}{2},\qquad \sigma = \frac{x_{\max}-x_{\min}}{2\sqrt{3}}.
\end{equation}
We apply this transform attribute-wise (i.e., per dimension) to descriptor vectors; in particular, we write $\hat{\mathbf{v}}_i$ and $\hat{\mathbf{u}}_j$ for the normalized versions of $\mathbf{v}_i$ and $\mathbf{u}_j$, respectively.
This choice corresponds to standardizing a uniform distribution.

\subsection{Base-Model and Recursive-Hypernetwork Structures}
\label{details-method-base-structure}

\paragraph{Overview.}
In our experiments, each generated base model is implemented as a stack of layers, where each layer composes several operations (e.g., a linear/conv operator together with optional normalization, activation, dropout \cite{srivastava2014dropout}, and shortcut \cite{he2016identity}).
In particular, whenever we say that UHN generates a model in our experiments, the generated parameters always instantiate one of the base-model families defined in this section, and that base model is
the composition of the layer primitives specified below.
As described in the main text, every scalar weight is represented by an \emph{index descriptor}.
Each layer is represented by a \emph{local structure descriptor}, and the full model is represented by a \emph{global structure descriptor} together with a \emph{task descriptor}.
We first define the layer types used throughout the paper, and then specify how these layers are composed into full models.

\subsubsection{Layer primitives}
\paragraph{Shared conventions.}
For graph and KAN layers (e.g., GCN \cite{kipf2016semi} / GAT \cite{velivckovic2017graph} / KAN \cite{liu2024kan}), let $X\in\mathbb{R}^{n\times d_{\mathrm{in}}}$ be the input (a batch of $n$ row vectors) and $Y\in\mathbb{R}^{n\times d_{\mathrm{out}}}$ its output.
For graph layers with $n$ nodes, let $A\in\{0,1\}^{n\times n}$ be the adjacency matrix and let $\tilde{A}=A+I$ be the adjacency matrix with added self-loops.
Let $\tilde{D}$ be the degree matrix of $\tilde{A}$.
We write $\mathcal{E}$ for the directed edge set induced by $\tilde{A}$.
For a node index $v\in\{1,\dots,n\}$, we write $\Gamma(v)=\{u\,\mid\,(u\to v)\in\mathcal{E}\}$ for the set of source neighbors of $v$.
For MHA \cite{vaswani2017attention} layers, we use $X\in\mathbb{R}^{n\times T\times d}$ and $Y\in\mathbb{R}^{n\times T\times d}$, where $T$ is the sequence length.
For embedding layers, let $V$ be the vocabulary size and use token indices $X\in\{1,\dots,V\}^{n\times T}$ as input and $Y\in\mathbb{R}^{n\times T\times d}$ as output.
For convolution layers, let $X\in\mathbb{R}^{n\times c_{\mathrm{in}}\times h\times w}$ and $Y\in\mathbb{R}^{n\times c_{\mathrm{out}}\times h'\times w'}$ (channels-first).
For linear layers, the input $X$ may be a matrix ($X\in\mathbb{R}^{n\times d_{\mathrm{in}}}$), a sequence tensor ($X\in\mathbb{R}^{n\times T\times d_{\mathrm{in}}}$), or an image tensor ($X\in\mathbb{R}^{n\times c_{\mathrm{in}}\times h\times w}$).
The output $Y$ is either a matrix ($Y\in\mathbb{R}^{n\times d_{\mathrm{out}}}$) or a sequence tensor ($Y\in\mathbb{R}^{n\times T\times d_{\mathrm{out}}}$).
We never produce a 4D output from a linear layer: for image inputs we always apply adaptive pooling to reduce $X$ to 2D before the linear map.

We write $\alpha(\cdot)$ for the chosen activation, $\mathcal{R}(\cdot)$ for dropout and $\mathcal{N}(\cdot)$ for the chosen (optional) normalization. When enabled, $\mathcal{N}$ corresponds to LayerNorm \cite{ba2016layer} for non-convolutional layers, and to GroupNorm \cite{wu2018group} for convolutional layers; otherwise, $\mathcal{N}$ corresponds to identity map.
We write $\langle \cdot,\cdot\rangle$ for the standard Euclidean inner product.

We optionally apply a deterministic pooling operator $\mathcal{P}(\cdot)$ to the layer input.
For linear layers, $\mathcal{P}$ is either the identity, reshaping input to a matrix, adaptive pooling (for image-like inputs), averaging over the sequence dimension or selection of the first token (for sequence inputs).

For convolution layers, we use either (i) explicit pooling, where $\mathcal{P}$ is identity/average/max pooling applied to the input, or (ii) stride-2 convolution (in this case we set $\mathcal{P}$ to identity and perform downsampling via the stride).

The shortcut branch is optional: when enabled, $\mathrm{Skip}(X)=X$, and when disabled we set $\mathrm{Skip}(X)=0$.

When adding a bias vector to a batched tensor, broadcasting is implied.

\paragraph{Linear layer ($f_{\mathrm{lin}}$).}
A linear layer with bias computes
\begin{equation}
    Y = \mathrm{Skip}(\mathcal{P}(X)) + \mathcal{R}\Big( \mathcal{N}(\alpha(\mathcal{P}(X))) W + b \Big),
\end{equation}
where $W\in\mathbb{R}^{d_{\mathrm{in}}\times d_{\mathrm{out}}}$ and $b\in\mathbb{R}^{d_{\mathrm{out}}}$.

\begin{itemize}
    \item Generated parameters: $W$, $b$.
    \item Active index descriptor fields: layer\_idx, layer\_type, param\_type, out\_idx, in\_idx.
    \item Active local structure descriptor fields: layer\_idx, layer\_type, bias\_type, norm\_type, shortcut\_type, output\_size, input\_size, activation\_type, activation\_param, dropout\_rate, input\_pooling\_reshape\_type, initialization\_type.
\end{itemize}

\paragraph{Convolution layer ($f_{\mathrm{conv}}$).}
For a 2D convolution with kernel size $k\times k$, we compute
\begin{equation}
Y = \mathrm{Skip}(X) + \mathcal{R}\Big(\mathrm{Conv}\big(\mathcal{P}(\mathcal{N}(\alpha(X)));\, W\big) + b\Big).
\end{equation}
where $W\in\mathbb{R}^{c_{\mathrm{out}}\times c_{\mathrm{in}}\times k\times k}$ and $b\in\mathbb{R}^{c_{\mathrm{out}}}$.

\begin{itemize}
    \item Generated parameters: $W$, $b$.
    \item Active index descriptor fields: layer\_idx, layer\_type, param\_type, out\_idx, in\_idx, kernel\_h\_idx, kernel\_w\_idx.
    \item Active local structure descriptor fields: layer\_idx, layer\_type, bias\_type, norm\_type, shortcut\_type, output\_size, input\_size, activation\_type, activation\_param, dropout\_rate, group\_num, kernel\_size, stage\_wise\_pooling\_type, initialization\_type.
\end{itemize}

\paragraph{GCN layer ($f_{\mathrm{gcn},A}$).}
A GCN layer with bias computes
\begin{equation}
    Y = \tilde{D}^{-\frac12}\tilde{A}\tilde{D}^{-\frac12}\,\Big(\mathcal{R}\big( \mathcal{N}(\alpha(X)) \big) W\Big) + b,
\end{equation}
where $W\in\mathbb{R}^{d_{\mathrm{in}}\times d_{\mathrm{out}}}$ and $b\in\mathbb{R}^{d_{\mathrm{out}}}$.

\begin{itemize}
    \item Generated parameters: $W$, $b$.
    \item Active index descriptor fields: layer\_idx, layer\_type, param\_type, out\_idx, in\_idx.
    \item Active local structure descriptor fields: layer\_idx, layer\_type, bias\_type, norm\_type, output\_size, input\_size, activation\_type, activation\_param, dropout\_rate, initialization\_type.
\end{itemize}

\paragraph{GAT layer ($f_{\mathrm{gat},A}$).}
A GAT layer with bias applies a linear projection $W\in\mathbb{R}^{d_{\mathrm{in}}\times d_{\mathrm{out}}}$ and reshapes into $h$ heads,
\begin{equation}
    Z=\mathcal{R}\big( \mathcal{N}(\alpha(X)) \big)W,\qquad Z^{(r)}_v\in\mathbb{R}^{d_h},\; r\in\{1,\dots,h\}.
\end{equation}
where $d_h = d_{\mathrm{out}}/h$ denote the per-head output dimension.
For each directed edge $(u\to v)\in\mathcal{E}$ and head $r$, with attention vectors $a_{\mathrm{src}},a_{\mathrm{dst}}\in\mathbb{R}^{h\times d_h}$,
\begin{equation}
\begin{aligned}
    e^{(r)}_{u\to v} &= \mathrm{LeakyReLU}_{0.2}\Big( \langle a^{(r)}_{\mathrm{src}}, Z^{(r)}_u\rangle + \langle a^{(r)}_{\mathrm{dst}}, Z^{(r)}_v\rangle \Big),\\
    \beta^{(r)}_{u\to v} &= \frac{\exp\big(e^{(r)}_{u\to v}\big)}{\sum_{u'\in\Gamma(v)}\exp\big(e^{(r)}_{u'\to v}\big)}.
\end{aligned}
\end{equation}
The head output is
\begin{equation}
    Y^{(r)}_v = \sum_{u\in\Gamma(v)} \beta^{(r)}_{u\to v}\, Z^{(r)}_u,
\end{equation}
We combine heads by concatenation or averaging and add bias,
\begin{equation}
    Y_v =
    \begin{cases}
        \mathrm{Concat}\big(Y^{(1)}_v,\dots,Y^{(h)}_v\big)+b, & \text{(concat)},\\
        \frac{1}{h}\sum_{r=1}^h Y^{(r)}_v+b, & \text{(avg)}.
    \end{cases}
\end{equation}
where $\mathrm{Concat}(\cdot)$ denotes concatenation across heads and $b\in\mathbb{R}^{d_{\mathrm{out}}}$ (concat) or $b\in\mathbb{R}^{d_h}$ (avg).

\begin{itemize}
    \item Generated parameters: $W$, $b$, $a_{\mathrm{src}}$, $a_{\mathrm{dst}}$.
    \item Active index descriptor fields: layer\_idx, layer\_type, param\_type, out\_idx, in\_idx.
    \item Active local structure descriptor fields: layer\_idx, layer\_type, bias\_type, norm\_type, output\_size, input\_size, activation\_type, activation\_param, dropout\_rate, num\_heads, head\_concat\_type, initialization\_type.
\end{itemize}

\paragraph{Embedding layer ($f_{\mathrm{emb}}$).}
An embedding layer returns
\begin{equation}
    Y = \mathcal{R}\!\left(\mathrm{Emb}(X) + P_{1:T}\right).
\end{equation}
Here $\mathrm{Emb}(X)\in\mathbb{R}^{n\times T\times d}$ denotes embedding lookup in $E$ and $P_{1:T}\in\mathbb{R}^{T\times d}$ is broadcast over the batch dimension,
where $E\in\mathbb{R}^{V\times d}$ is the token embedding table and $P\in\mathbb{R}^{T_{\max}\times d}$ the positional embedding table, with $T_{\max}$ the maximum supported sequence length.

\begin{itemize}
    \item Generated parameters: $E$, $P$.
    \item Active index descriptor fields: layer\_idx, layer\_type, param\_type, out\_idx, embedding\_idx, sequence\_idx.
    \item Active local structure descriptor fields: layer\_idx, layer\_type, output\_size (embedding dim), dropout\_rate, embedding\_num, max\_sequence\_length, initialization\_type.
\end{itemize}

\paragraph{Multi-head attention (MHA) layer ($f_{\mathrm{mha}}$).}
An MHA layer with $h$ heads (head dimension $d_h=d/h$) first applies a pre-activation formulation:
\begin{equation}
    \bar{X}=\mathcal{N}(\alpha(X)).
\end{equation}
We then compute projections
\begin{equation}
    Q=\bar{X}W_Q+b_Q,\qquad K=\bar{X}W_K+b_K,\qquad V=\bar{X}W_V+b_V,
\end{equation}
reshape $Q,K,V$ into $h$ heads (each in $\mathbb{R}^{n\times T\times d_h}$), and apply scaled dot-product attention per head:
\begin{equation}
    A^{(r)}=\mathcal{R}\!\left(\mathrm{softmax}\Big(\tfrac{Q^{(r)}(K^{(r)})^{\top}}{\sqrt{d_h}}\Big)\right)V^{(r)},\qquad r\in\{1,\dots,h\}.
\end{equation}
We concatenate head outputs $\mathrm{Concat}(A^{(1)},\dots,A^{(h)})\in\mathbb{R}^{n\times T\times d}$ and apply the output projection,
\begin{equation}
    Y=\mathrm{Skip}(X)+\mathcal{R}\Big(\mathrm{Concat}(A^{(1)},\dots,A^{(h)})W_O+b_O\Big).
\end{equation}

\begin{itemize}
    \item Generated parameters: $W_Q,W_K,W_V,b_Q,b_K,b_V,W_O,b_O$, where $W_Q,W_K,W_V,W_O\in\mathbb{R}^{d\times d}$ and $b_Q,b_K,b_V,b_O\in\mathbb{R}^{d}$.
    \item Active index descriptor fields: layer\_idx, layer\_type, param\_type, out\_idx, in\_idx.
    \item Active local structure descriptor fields: layer\_idx, layer\_type, bias\_type, norm\_type, shortcut\_type, output\_size (embedding dim), input\_size (also set to embedding dim), activation\_type, activation\_param, dropout\_rate, num\_heads, initialization\_type.
\end{itemize}

\paragraph{KAN layer ($f_{\mathrm{kan}}$).}
A KAN layer combines a base linear map and a spline expansion. We compute
\begin{equation}
    Y = \alpha(X) W_{\mathrm{base}} + b_{\mathrm{base}} + \Phi(X;\,g_{\min},\Delta,\kappa, p) \widetilde{W}_{\mathrm{spline}},
\end{equation}
where $W_{\mathrm{base}}\in\mathbb{R}^{d_{\mathrm{in}}\times d_{\mathrm{out}}}$ and $b_{\mathrm{base}}\in\mathbb{R}^{d_{\mathrm{out}}}$.
Concretely, let the spline weights be a tensor $W_{\mathrm{spline}}\in\mathbb{R}^{d_{\mathrm{in}}\times (G+p)\times d_{\mathrm{out}}}$ and let spline scales be $S\in\mathbb{R}^{d_{\mathrm{in}}\times d_{\mathrm{out}}}$, where $G$ is the grid size and $p$ is the spline order.
We form the scaled spline weight matrix
\begin{equation}
    \widetilde{W}_{\mathrm{spline}} = \mathrm{reshape}\big( W_{\mathrm{spline}} \odot S,\; (d_{\mathrm{in}}(G+p),\, d_{\mathrm{out}})\big),
\end{equation}
where the elementwise multiplication $W_{\mathrm{spline}}\odot S$ uses broadcasting of $S$ along the grid dimension (of size $G+p$).
The basis map $\Phi(\cdot)$ stacks B-spline basis functions of order $p$; for the above batch input $X\in\mathbb{R}^{n\times d_{\mathrm{in}}}$, it returns $\Phi(X;\,g_{\min},\Delta,\kappa,p)\in\mathbb{R}^{n\times d_{\mathrm{in}}(G+p)}$.
We parameterize the knot grid via trainable parameters $g_{\min}\in\mathbb{R}^{d_{\mathrm{in}}}$, $\Delta\in\mathbb{R}^{d_{\mathrm{in}}}$, and $\kappa\in\mathbb{R}^{d_{\mathrm{in}}\times(G+2\,p+1)}$ and set
\begin{equation}
    g = g_{\min} + \exp(\Delta)\odot \mathrm{cumsum}(\mathrm{softmax}(\kappa)),
\end{equation}
where $g$ is the monotonically increasing knot sequence (with $\exp(\Delta)$ broadcast along the grid axis for the elementwise product) and $\mathrm{cumsum}$ denotes cumulative summation along the grid axis.

\begin{itemize}
    \item Generated parameters: $W_{\mathrm{base}}$, $b_{\mathrm{base}}$, $W_{\mathrm{spline}}$, $S$, $g_{\min}$, $\Delta$, $\kappa$.
    \item Active index descriptor fields: layer\_idx, layer\_type, param\_type, out\_idx, in\_idx, grid\_idx.
    \item Active local structure descriptor fields: layer\_idx, layer\_type, bias\_type, output\_size, input\_size, activation\_type (base), activation\_param, grid\_size, spline\_order, initialization\_type.
\end{itemize}

\subsubsection{Model families}
\label{model-families}
\paragraph{Model definition.}
Let the base model be the composition of $L$ layers (functions) $\{f_j\}_{j=1}^L$:
\begin{equation}
    f(X)=f_L\circ f_{L-1}\circ\cdots\circ f_1(X).
\end{equation}
In our experiments, generating a base model means generating all trainable parameters that appear in these constituent layers.
We represent each base model with a \emph{global structure descriptor} and a \emph{task descriptor}.
The shared global structure descriptor fields (active for all models) are model\_type and num\_layers.
Additional global structure descriptor fields are activated depending on the model family: CNNs use cnn\_stage\_num, Transformers \cite{vaswani2017attention} use num\_encoders, and recursive hypernetworks use num\_structure\_freqs and num\_index\_freqs.
The task descriptor is shared across all models and includes task\_type and dataset\_type.
Below we summarize the model families used in our experiments in a compact mathematical form.

\paragraph{MLP / CNN model.}
A MLP/CNN is a stack of convolutional layers followed by linear layers
\begin{equation}
    f(X)=f^{(J)}_{\mathrm{lin}}\circ\cdots\circ f^{(1)}_{\mathrm{lin}}\big( f^{(K)}_{\mathrm{conv}}\circ\cdots\circ f^{(1)}_{\mathrm{conv}}(X)\big),
\end{equation}
where $K$ is the number of conv layers and $J$ is the number of linear layers (including the output layer). $f^{(1)}_{\mathrm{lin}}$ includes reshaping/pooling needed to map the its input to a matrix.
When $K=0$, the model reduces to an MLP.

\paragraph{GCN model.}
A GCN is a composition of graph convolution layers
\begin{equation}
    f(X,A)=f^{(L)}_{\mathrm{gcn},A}\circ\cdots\circ f^{(1)}_{\mathrm{gcn},A}(X),
\end{equation}

\paragraph{GAT model.}
A GAT is a composition of graph attention layers
\begin{equation}
    f(X,A)=f^{(L)}_{\mathrm{gat},A}\circ\cdots\circ f^{(1)}_{\mathrm{gat},A}(X),
\end{equation}

\paragraph{Transformer model.}
A Transformer is a composition of an embedding layer, a stack of $J$ encoder blocks, and an output linear layer.
Let $\mathrm{Enc}^{(j)}(\cdot)$ denote the $j$th Transformer encoder block (one MHA layer + $K_{\mathrm{j}}$ linear layers)
\begin{equation}
    \mathrm{Enc}^{(j)}(X)=f^{(K_{\mathrm{j}})}_{\mathrm{lin}}\circ\cdots\circ f^{(1)}_{\mathrm{lin}}\circ f_{\mathrm{mha}}(X).
\end{equation}
Then
\begin{equation}
    f(X) = f_{\mathrm{lin}}\Big( \mathrm{Enc}^{(J)}\circ\cdots\circ\mathrm{Enc}^{(1)}(f_{\mathrm{emb}}(X)) \Big),
\end{equation}
where the output linear layer $f_{\mathrm{lin}}$ includes selecting the first token representation.

\paragraph{KAN model.}
A KAN model is a composition of KAN layers
\begin{equation}
    f(X)=f^{(L)}_{\mathrm{kan}}\circ\cdots\circ f^{(1)}_{\mathrm{kan}}(X),
\end{equation}

\subsubsection{Recursive hypernetwork}
\label{recursive-hypernetwork-model}
A recursive hypernetwork (i.e., a generated hypernetwork) is represented as a conditional mapping
\begin{equation}
    H\big(\mathbf{v}_{i},\{\mathbf{u}_j\}_{j=1}^{L'}\big)
    = f_{\mathrm{lin}}\Big(h_{\mathbf{v}}\big(\gamma_{\mathbf{B}_{\mathbf{v}}}(\hat{\mathbf{v}}_{i})\big)
    + h_{\mathbf{u}}\big(\{\gamma_{\mathbf{B}_{\mathbf{u}}}(\hat{\mathbf{u}}_j)\}_{j=1}^{L'}\big)\Big),
\end{equation}
where $\mathbf{v}_i$ is the index descriptor, $\mathbf{u}_j=[\mathbf{s}_g;\mathbf{t};\mathbf{s}_{\ell,j}]$ are $L'$ per-layer task-structure descriptors, and $\hat{\mathbf{v}}_{i}$ and $\hat{\mathbf{u}}_j$ denote the corresponding attribute-wise normalized descriptors.
Here $\gamma_{\mathbf{B}}(\cdot)$ denotes the fixed Gaussian Fourier feature map \cite{tancik2020fourier}, with fixed matrices $\mathbf{B}_{\mathbf{v}}$ and $\mathbf{B}_{\mathbf{u}}$.
We write $\boldsymbol{\phi}_i:=\gamma_{\mathbf{B}_{\mathbf{v}}}(\hat{\mathbf{v}}_{i})$ for the index Fourier features and $\boldsymbol{\psi}_j:=\gamma_{\mathbf{B}_{\mathbf{u}}}(\hat{\mathbf{u}}_j)$ for the per-layer token Fourier features.

For generality, we write the index branch $h_{\mathbf{v}}$ as a composition of $K$ linear layers,
\begin{equation}
    h_{\mathbf{v}}(\boldsymbol{\phi}_i) = f_{\mathrm{lin},\mathbf{v}}^{(K)}\circ \cdots \circ f_{\mathrm{lin},\mathbf{v}}^{(1)}(\boldsymbol{\phi}_i),
\end{equation}

Similarly, we write the task-structure branch $h_{\mathbf{u}}$ as a MHA layer followed by $J$ linear layers,
\begin{equation}
    h_{\mathbf{u}}\big(\{\boldsymbol{\psi}_j\}_{j=1}^{L'}\big)
    = f_{\mathrm{lin},\mathbf{u}}^{(J)}\circ \cdots \circ f_{\mathrm{lin},\mathbf{u}}^{(1)}\circ f_{\mathrm{mha}}\big(\{\boldsymbol{\psi}_j\}_{j=1}^{L'}\big),
\end{equation}
where one of the linear layers in the task-structure branch includes averaging over the token (sequence) dimension.
Finally, the fused representation is passed through the output linear layer $f_{\mathrm{lin}}$ to predict the weight scalar.

In the recursive setting, we reuse the same descriptor normalization statistics and Fourier matrices as the root UHN.

\subsection{Initialization target statistics}
\label{details-method-init-targets}
We define an initialization target for each generated \emph{parameter group} $g$ (i.e., each layer \emph{component} such as a weight matrix, bias vector, or attention projection \cite{vaswani2017attention, velivckovic2017graph}) in terms of a desired mean and standard deviation $(\mu^*(g),\sigma^*(g))$.
During the initialization phase, we match the parameter statistics $(\mu(g),\sigma(g))$ of the generated component $g$ to this target in order to stabilize optimization of the main training phase.
Throughout this subsection, $\mathcal{U}(a,b)$ denotes the continuous uniform distribution on $[a,b]$, and $\mathcal{N}(\mu,\sigma^2)$ denotes a Gaussian distribution with mean $\mu$ and variance $\sigma^2$.

\paragraph{Default vs. zero targets.}
We support two initialization modes: (i) \textsc{Default}, where targets mirror common library initializations \cite{paszke2019pytorch, fey2019fast}, and (ii) \textsc{Zero}, where $(\mu^*,\sigma^*)=(0,0)$ for all parameter groups.
If not specified otherwise, we use \textsc{Default}.

\paragraph{Default target statistics by layer.}
For \textsc{Default}, we use the following targets.
For Linear, convolution, embedding, and MHA \cite{vaswani2017attention} layers, these targets are chosen to match PyTorch default initializations \cite{paszke2019pytorch}.
For GCN \cite{kipf2016semi} and GAT \cite{velivckovic2017graph} layers, we match the defaults used by PyTorch Geometric \cite{fey2019fast}.
For KAN layers, we follow the original KAN paper with several modifications \cite{liu2024kan}, as described below.
\begin{itemize}
    \item \textbf{Linear layer.} For weights and biases we use $\mu^*=0$ and $\sigma^*=\frac{1}{\sqrt{3\,d_{\mathrm{in}}}}$, where $d_{\mathrm{in}}$ is the input dimension.
    \item \textbf{Convolution layer.} For a $k\times k$ convolution with $c_{\mathrm{in}}$ input channels we use $\mu^*=0$ and $\sigma^*=\frac{1}{\sqrt{3\,c_{\mathrm{in}}k^2}}$ (for both weights and biases).
    \item \textbf{GCN layer.} For weights we use $\mu^*=0$ and $\sigma^*=\sqrt{\frac{2}{d_{\mathrm{in}}+d_{\mathrm{out}}}}$, where $d_{\mathrm{in}}$ and $d_{\mathrm{out}}$ are the input/output dimensions;
    biases are set to zero, i.e., $(\mu^*_{\mathrm{b}},\sigma^*_{\mathrm{b}})=(0,0)$.
    \item \textbf{GAT layer.} For the main weights we use
    $\mu^*=0$ and $\sigma^*=\sqrt{\frac{2}{d_{\mathrm{in}}+h\,d_{\mathrm{out}}}}$, where $d_{\mathrm{in}}$ is the input dimension, $d_{\mathrm{out}}$ is the output dimension per head, and $h$ is the number of heads;
    biases are set to zero.
    For attention parameters (source and target) we use
    $\mu^*=0$ and $\sigma^*=\sqrt{\frac{2}{d_{\mathrm{out}}+h}}$.
    \item \textbf{Embedding layer.} Embedding weights use $(\mu^*,\sigma^*)=(0,1)$, while positional embeddings use $(0,0)$.
    \item \textbf{Multi-head attention (MHA).} For Q/K/V projection weights we use a Glorot-style \cite{glorot2010understanding} target
    $\mu^*=0$ and $\sigma^*=\sqrt{\frac{2}{d+d}}=\sqrt{\frac{1}{d}}$, where $d$ is the MHA dimension;
    all projection biases are set to zero.
    For the output projection weight we match $\mathcal{U}(-1/\sqrt{d},1/\sqrt{d})$, i.e., $\mu^*=0$ and $\sigma^*=\frac{1}{\sqrt{3d}}$, and set its bias to zero.
    \item \textbf{KAN layer.} Base weights and spline scales match $\mathcal{U}(-1/\sqrt{d_{\mathrm{in}}},1/\sqrt{d_{\mathrm{in}}})$ (thus $\mu^*=0$, $\sigma^*=1/\sqrt{3\,d_{\mathrm{in}}}$, with $d_{\mathrm{in}}$ the layer input dimension). Base biases are set to zero.
    Spline weights follow $\mathcal{N}(0,0.1^2)$.
    Grid parameters are initialized deterministically with zero variance: grid lower bounds are set to $-1$, grid lengths are set to $\log 2$, and grid knots are set to $0$.
\end{itemize}

\subsection{Hyperparameter selection}
\label{details-method-sweep}
Here we summarize the hyperparameters and sweep protocol used across experiments.
\paragraph{Architecture hyperparameters.}
The main model (architecture) hyperparameters are the number of Fourier-feature \cite{tancik2020fourier} frequencies $F_{\mathbf{v}}$ for index encoding, the Fourier-feature scale $\sigma$ (for both index encoding and structure/task encoding), the UHN hidden dimension $d$, and the number of UHN residual blocks $N_{\mathrm{blk}}$.
When used, the task-structure encoder introduces additional hyperparameters, including the number of Fourier-feature frequencies $F_{\mathbf{u}}$ for task-structure descriptors and the number of attention heads \cite{vaswani2017attention} $h$.
In all experiments, we fix $\sigma=100$, and (when using the task-structure encoder) we set $(F_{\mathbf{u}},h)=(32,4)$.

Unless otherwise specified, we use $(F_{\mathbf{v}}, d, N_{\mathrm{blk}})=(1024,64,2)$ for the single-model MNIST \cite{lecun1998gradient} experiment and for recursive experiments, and $(F_{\mathbf{v}}, d, N_{\mathrm{blk}})=(2048,128,2)$ for all other experiments.

\paragraph{Swept hyperparameters.}
For each experimental setting, we sweep the initialization hyperparameters $(S_{\mathrm{init}},\eta_{\mathrm{init}})$, where $S_{\mathrm{init}}$ denotes the number of initialization steps and $\eta_{\mathrm{init}}$ denotes the initialization learning rate.
For the training phase, we sweep either $(S_{\mathrm{train}},\eta_{\mathrm{train}})$ when using step-based training budgets (multi-task/recursive settings) or $(E_{\mathrm{train}},\eta_{\mathrm{train}})$ when using epoch-based training budgets (single-/multi-model settings), where $S_{\mathrm{train}}$ denotes the number of training steps, $E_{\mathrm{train}}$ denotes the number of training epochs, and $\eta_{\mathrm{train}}$ denotes the training learning rate.

\paragraph{Sweep protocol.}
To reduce search cost, we use a staged (rather than fully joint) sweep:
\begin{enumerate}
    \item Fix the training budget and sweep $\eta_{\mathrm{train}}$ with $S_{\mathrm{init}}=0$.
    \item Fix the selected training hyperparameters and sweep $(S_{\mathrm{init}},\eta_{\mathrm{init}})$.
    \item Fix the selected initialization hyperparameters and refine the training hyperparameters.
\end{enumerate}
We denote the fixed training budget in the first sweeping stage as $E_0$ (epochs) in single-/multi-model settings and as $S_0$ (steps) in multi-task/recursive settings.

\paragraph{Validation/selection criterion.}
In the single-model setting, validation is performed on the single target model.
For multi-model experiments, we select hyperparameters by average validation performance over architectures in $M_{\mathrm{val}}$.
For multi-task experiments, we compare configurations using a rank-based aggregate across tasks (i.e., Borda count \cite{borda1781m}) with task-specific guardrails.
The selection criterion for recursive experiments follows the multi-task protocol.

\section{Single-Model Experiments}
\label{details-exp-single-model}
This section provides implementation and evaluation details supporting the single-model experiments in \cref{single-model-exp}, including datasets, base model architectures, additional results and the hyperparameter selection.

\subsection{Datasets}
\textbf{MNIST.}
MNIST \cite{lecun1998gradient} is an image classification dataset with 10 classes, 60{,}000 training samples, and 10{,}000 test samples ($28\times 28$ grayscale).
During hyperparameter sweeps we reserve 20\% of the training set for validation; final results are obtained by training on the full training set.
We apply standard per-pixel normalization.

\textbf{CIFAR-10.}
CIFAR-10 \cite{krizhevsky2009learning} is an image classification dataset with 10 classes, 50{,}000 training samples, and 10{,}000 test samples ($32\times 32$ RGB).
During sweeps we reserve 10\% of the training set for validation.
For training we use augmentation following \cite{zagoruyko2016wide}: 4-pixel reflection padding, random horizontal flip, random $32\times 32$ crops, and per-channel normalization; for testing we use only normalization.

\textbf{Citation networks (Cora, CiteSeer, PubMed).}
Cora, CiteSeer, and PubMed \cite{sen2008collective} are citation-network benchmarks for graph node classification.
\Cref{tab:citation-network-stats} summarizes dataset statistics, as reported in \cite{yang2016revisiting}, where nodes correspond to documents, edges correspond to citations, features are bag-of-words vectors.
We use the standard splits with 140/500/1000 nodes for train/val/test on Cora, 120/500/1000 on CiteSeer, and 60/500/1000 on PubMed following \cite{kipf2016semi}.
We normalize node features for both training and evaluation.

\begin{table}[t]
\centering
\caption{Citation-network dataset statistics, as reported in \cite{yang2016revisiting}, used in our experiments.}
\label{tab:citation-network-stats}
\scriptsize
\begin{tabular}{lccc}
\toprule
 & Cora & CiteSeer & PubMed \\
\midrule
Nodes & 2708 & 3327 & 19717 \\
Edges & 5429 & 4732 & 44338 \\
Features & 1433 & 3703 & 500 \\
Classes & 7 & 6 & 3 \\
\bottomrule
\end{tabular}
\end{table}

\textbf{AG News.}
AG News \cite{zhang2015character} is a text classification dataset with 4 classes, 120{,}000 training samples, and 7{,}600 test samples.
During sweeps we reserve 20\% of the training set for validation.
We use a simple English tokenizer (lowercasing + regex tokenization), build a vocabulary of the $5{,}000$ most frequent tokens, including \texttt{<CLS>} (classification token), \texttt{<PAD>} (padding), \texttt{<UNK>} (unknown), and use a maximum sequence length of 128 (including the first \texttt{<CLS>} token).
We apply dynamic padding per batch and use a padding mask.

\textbf{IMDB.}
IMDB \cite{maas2011learning} is a binary text classification dataset with 25{,}000 training samples and 25{,}000 test samples.
During sweeps we reserve 20\% of the training set for validation.
We use the same tokenizer and vocabulary construction as AG News, but use a maximum sequence length of 512.

\textbf{Formula regression (special functions).}
We evaluate formula regression on 15 special functions (following the KAN benchmark \cite{liu2024kan}; listed in \cref{tab:single-model-universality-formula}).
For each function, we sample 1{,}000 training and 1{,}000 test inputs uniformly over $(x_1,x_2)\in[0,1)^2$ (except for special functions yv and kv, where we sample $x_2\in[\epsilon,1)$ to avoid numerical instabilities; we use $\epsilon=0.1$).
For special function ellipj the target is $\mathbb{R}^4$; otherwise it is scalar.
We reserve 20\% of the training set for validation, use full-batch training, and normalize inputs to $[-1,1)$.

\subsection{Base-model architectures}
\label{details-base-model-architectures}
The following base models are instances of the model families in \cref{model-families}; their parameters are generated by UHN.
All base-model input/output dimensions are adapted to the corresponding dataset.
\begin{itemize}
    \item \textbf{MLP (MNIST).} Two hidden layers of width 128; bias enabled; activation: LeakyReLU (slope 0.1); normalization: LayerNorm \cite{ba2016layer}; no dropout \cite{srivastava2014dropout}; residual shortcuts \cite{he2016identity} enabled.
    \item \textbf{CNN-8 (MNIST).} Four-stage CNN; layers per stage 1/2/2/2; widths 16/16/32/64; bias enabled; kernel size $3$; downsampling (via stride-2 convolution) in the third and fourth stages; GroupNorm \cite{wu2018group} with 4 groups; average pooling after convolutions; LeakyReLU (0.1); no dropout; residual shortcuts enabled; classifier is a final linear layer (overall, a lightweight pre-activation ResNet-style \cite{he2016identity} variant).
    \item \textbf{CNN-20/32/44/56 (CIFAR-10).} Same as CNN-8 (MNIST), but with layers per stage 1/$n$/$n$/$n$ and $n\in\{6,10,14,18\}$.
    \item \textbf{GCN \cite{kipf2016semi}.} One hidden layer of width 64; bias enabled; activation: LeakyReLU (0.1); dropout 0.5; no normalization.
    \item \textbf{GAT \cite{velivckovic2017graph}.} One hidden layer with 8 heads and head dimension 8; 8 output heads for PubMed and 1 otherwise; bias enabled; activation: ELU \cite{clevert2015fast}; dropout 0.6; no normalization.
    \item \textbf{Transformer-2L \cite{vaswani2017attention} (AG News).} Two encoder blocks; 2 attention heads; 2 feed-forward layers per encoder; embedding dimension 64; bias and residual shortcuts enabled; normalization: LayerNorm; activation: LeakyReLU (0.1); dropout 0.2.
    \item \textbf{Transformer-1L (IMDB).} Same as Transformer-2L (AG News), but with 1 encoder block and dropout 0.4.
    \item \textbf{KAN-g5/g10 \cite{liu2024kan}.} One hidden layer of width 5; grid size 5 by default (KAN-g5) and 10 for special functions jv and yv (KAN-g10); spline order 3; bias enabled; base activation: SiLU \cite{hendrycks2016gaussian}.
\end{itemize}

\subsection{Formula-regression results (universality experiment)}
\Cref{tab:single-model-universality-formula} reports the full per-function formula-regression results for the single-model universality experiment, as a continuation of \cref{sec:results-single-task}.
\begin{table}[t]
\centering
\caption{Single-model universality results for formula regression on the 15 special functions following the KAN benchmark \cite{liu2024kan}. Model (\#Params) reports the base model architecture and its trainable parameter count; the trainable parameter count of UHN is fixed at 612{,}117 for all functions. RMSE denotes test root-mean-squared error (lower is better).}
\label{tab:single-model-universality-formula}
\scriptsize
\begin{tabular}{ll ll}
\toprule
Function & Model (\#Params) & RMSE (Direct) & RMSE (UHN) \\
\midrule
Jacobian elliptic functions (ellipj) & KAN-g5 (407) & $6.78\times 10^{-4} \pm 1.32\times 10^{-4}$ & $6.44\times 10^{-4} \pm 5.20\times 10^{-5}$ \\
Incomplete elliptic integral of the first kind (ellipkinc) & KAN-g5 (254) & $5.37\times 10^{-4} \pm 1.40\times 10^{-4}$ & $5.99\times 10^{-4} \pm 9.32\times 10^{-5}$ \\
Incomplete elliptic integral of the second kind (ellipeinc) & KAN-g5 (254) & $6.02\times 10^{-4} \pm 2.58\times 10^{-4}$ & $5.23\times 10^{-4} \pm 4.51\times 10^{-5}$ \\
Bessel function of the first kind (jv) & KAN-g10 (364) & $1.80\times 10^{-3} \pm 8.04\times 10^{-4}$ & $2.32\times 10^{-3} \pm 1.12\times 10^{-3}$ \\
Bessel function of the second kind (yv) & KAN-g10 (364) & $1.09\times 10^{-2} \pm 4.05\times 10^{-3}$ & $8.19\times 10^{-3} \pm 3.67\times 10^{-3}$ \\
Modified Bessel function of the second kind (kv) & KAN-g5 (254) & $2.11\times 10^{-2} \pm 1.26\times 10^{-2}$ & $1.04\times 10^{-2} \pm 4.77\times 10^{-3}$ \\
Modified Bessel function of the first kind (iv) & KAN-g5 (254) & $2.94\times 10^{-3} \pm 1.74\times 10^{-3}$ & $2.91\times 10^{-3} \pm 9.40\times 10^{-4}$ \\
Associated Legendre function ($m=0$) (lpmv0) & KAN-g5 (254) & $7.06\times 10^{-4} \pm 1.57\times 10^{-4}$ & $7.28\times 10^{-4} \pm 6.04\times 10^{-5}$ \\
Associated Legendre function ($m=1$) (lpmv1) & KAN-g5 (254) & $2.22\times 10^{-3} \pm 1.42\times 10^{-4}$ & $1.57\times 10^{-3} \pm 3.48\times 10^{-4}$ \\
Associated Legendre function ($m=2$) (lpmv2) & KAN-g5 (254) & $2.74\times 10^{-4} \pm 2.95\times 10^{-5}$ & $9.61\times 10^{-4} \pm 1.16\times 10^{-4}$ \\
Spherical harmonics ($m=0$, $n=1$) (sph\_harm01) & KAN-g5 (254) & $1.62\times 10^{-4} \pm 2.66\times 10^{-5}$ & $3.70\times 10^{-4} \pm 1.64\times 10^{-4}$ \\
Spherical harmonics ($m=1$, $n=1$) (sph\_harm11) & KAN-g5 (254) & $2.11\times 10^{-4} \pm 4.92\times 10^{-5}$ & $4.52\times 10^{-4} \pm 1.79\times 10^{-4}$ \\
Spherical harmonics ($m=0$, $n=2$) (sph\_harm02) & KAN-g5 (254) & $2.22\times 10^{-4} \pm 9.92\times 10^{-5}$ & $4.40\times 10^{-4} \pm 1.85\times 10^{-4}$ \\
Spherical harmonics ($m=1$, $n=2$) (sph\_harm12) & KAN-g5 (254) & $2.44\times 10^{-4} \pm 1.34\times 10^{-5}$ & $5.44\times 10^{-4} \pm 1.63\times 10^{-4}$ \\
Spherical harmonics ($m=2$, $n=2$) (sph\_harm22) & KAN-g5 (254) & $1.54\times 10^{-4} \pm 4.90\times 10^{-6}$ & $5.33\times 10^{-4} \pm 8.24\times 10^{-5}$ \\
\bottomrule
\end{tabular}
\end{table}

\subsection{Hyperparameter selection}
\subsubsection{Universality experiment}
For UHN, we select all optimization hyperparameters by grid search, following the sweep protocol described in \cref{details-method-sweep}.
Concretely, we sweep (i) the initialization-stage learning rate and steps $(\eta_{\mathrm{init}}, S_{\mathrm{init}})$ and (ii) the training-stage learning rate and epochs $(\eta_{\mathrm{train}}, E_{\mathrm{train}})$.
\Cref{tab:sweep-grid} summarizes the sweep grids, and \cref{tab:sweep-best} reports the best hyperparameters selected by the grid search.

\begin{table}[t]
\centering
\caption{UHN hyperparameter sweep grids (universality experiment). Table entries list the sweep-grid values for $\eta_{\mathrm{init}}$, $S_{\mathrm{init}}$, $\eta_{\mathrm{train}}$, and $E_{\mathrm{train}}$; we use ``All'' when the same grid applies to all datasets/models within that task.}
\label{tab:sweep-grid}
\scriptsize
\begin{tabular}{lllccccc}
\toprule
Task & Dataset & Model & $E_0$ & $\eta_{\mathrm{init}}$ & $S_{\mathrm{init}}$ & $\eta_{\mathrm{train}}$ & $E_{\mathrm{train}}$ \\
\midrule
Image & MNIST & MLP & 100 & $\{5\mathrm{e}{-5},1\mathrm{e}{-4},2\mathrm{e}{-4}\}$ & $\{50,100,200\}$ & $\{5\mathrm{e}{-5},1\mathrm{e}{-4},2\mathrm{e}{-4}\}$ & $\{50,100,200\}$ \\
Image & MNIST & CNN-8 & 50 & $\{5\mathrm{e}{-5},1\mathrm{e}{-4},2\mathrm{e}{-4}\}$ & $\{50,100,200\}$ & $\{5\mathrm{e}{-5},1\mathrm{e}{-4},2\mathrm{e}{-4}\}$ & $\{30,50,100\}$ \\
Image & CIFAR-10 & CNN-20 & 200 & $\{5\mathrm{e}{-5},1\mathrm{e}{-4},2\mathrm{e}{-4}\}$ & $\{50,100,200\}$ & $\{5\mathrm{e}{-5},1\mathrm{e}{-4},2\mathrm{e}{-4}\}$ & $\{200,400,800\}$ \\
Text & AG News & Transformer-2L & 50 & $\{5\mathrm{e}{-5},1\mathrm{e}{-4},2\mathrm{e}{-4}\}$ & $\{50,100,200\}$ & $\{5\mathrm{e}{-5},1\mathrm{e}{-4},2\mathrm{e}{-4}\}$ & $\{30,50,100\}$ \\
Text & IMDB & Transformer-1L & 100 & $\{5\mathrm{e}{-5},1\mathrm{e}{-4},2\mathrm{e}{-4}\}$ & $\{50,100,200\}$ & $\{5\mathrm{e}{-5},1\mathrm{e}{-4},2\mathrm{e}{-4}\}$ & $\{50,100,200\}$ \\
Graph & All & All & 200 & $\{5\mathrm{e}{-5},1\mathrm{e}{-4},2\mathrm{e}{-4}\}$ & $\{50,100,200\}$ & $\{2\mathrm{e}{-5},5\mathrm{e}{-5},1\mathrm{e}{-4}\}$ & $\{100,200,400\}$ \\
Formula & All & All & 4000 & $\{5\mathrm{e}{-5},1\mathrm{e}{-4},2\mathrm{e}{-4}\}$ & $\{50,100,200\}$ & $\{2\mathrm{e}{-5},5\mathrm{e}{-5},1\mathrm{e}{-4}\}$ & $\{4000\}$ \\
\bottomrule
\end{tabular}
\end{table}

\begin{table}[t]
\centering
\caption{UHN best hyperparameters selected by grid search (universality experiment).}
\label{tab:sweep-best}
\scriptsize
\begin{tabular}{lllcccc}
\toprule
Task & Dataset & Model & $\eta_{\mathrm{init}}$ & $S_{\mathrm{init}}$ & $\eta_{\mathrm{train}}$ & $E_{\mathrm{train}}$ \\
\midrule
Image & MNIST & MLP & $2\mathrm{e}{-4}$ & 100 & $2\mathrm{e}{-4}$ & 100 \\
Image & MNIST & CNN-8 & $2\mathrm{e}{-4}$ & 200 & $1\mathrm{e}{-4}$ & 100 \\
Image & CIFAR-10 & CNN-20 & 0 & 0 & $2\mathrm{e}{-4}$ & 800 \\
\midrule
Graph & Cora & GCN & $1\mathrm{e}{-4}$ & 200 & $2\mathrm{e}{-5}$ & 400 \\
Graph & CiteSeer & GCN & $1\mathrm{e}{-4}$ & 50 & $1\mathrm{e}{-4}$ & 200 \\
Graph & PubMed & GCN & $1\mathrm{e}{-4}$ & 200 & $5\mathrm{e}{-5}$ & 200 \\
Graph & Cora & GAT & $2\mathrm{e}{-4}$ & 100 & $5\mathrm{e}{-5}$ & 400 \\
Graph & CiteSeer & GAT & $2\mathrm{e}{-4}$ & 200 & $5\mathrm{e}{-5}$ & 400 \\
Graph & PubMed & GAT & $2\mathrm{e}{-4}$ & 100 & $2\mathrm{e}{-5}$ & 400 \\
\midrule
Text & AG News & Transformer-2L & 0 & 0 & $5\mathrm{e}{-5}$ & 50 \\
Text & IMDB & Transformer-1L & $5\mathrm{e}{-5}$ & 100 & $1\mathrm{e}{-4}$ & 100 \\
\midrule
Formula & ellipj & KAN-g5 & $1\mathrm{e}{-4}$ & 200 & $1\mathrm{e}{-4}$ & 4000 \\
Formula & ellipkinc & KAN-g5 & $2\mathrm{e}{-4}$ & 50 & $1\mathrm{e}{-4}$ & 4000 \\
Formula & ellipeinc & KAN-g5 & $1\mathrm{e}{-4}$ & 50 & $1\mathrm{e}{-4}$ & 4000 \\
Formula & jv & KAN-g10 & $2\mathrm{e}{-4}$ & 200 & $5\mathrm{e}{-5}$ & 4000 \\
Formula & yv & KAN-g10 & $2\mathrm{e}{-4}$ & 100 & $1\mathrm{e}{-4}$ & 4000 \\
Formula & kv & KAN-g5 & $2\mathrm{e}{-4}$ & 200 & $1\mathrm{e}{-4}$ & 4000 \\
Formula & iv & KAN-g5 & $5\mathrm{e}{-5}$ & 100 & $1\mathrm{e}{-4}$ & 4000 \\
Formula & lpmv0 & KAN-g5 & $5\mathrm{e}{-5}$ & 200 & $5\mathrm{e}{-5}$ & 4000 \\
Formula & lpmv1 & KAN-g5 & $5\mathrm{e}{-5}$ & 200 & $5\mathrm{e}{-5}$ & 4000 \\
Formula & lpmv2 & KAN-g5 & $2\mathrm{e}{-4}$ & 200 & $1\mathrm{e}{-4}$ & 4000 \\
Formula & sph\_harm01 & KAN-g5 & 0 & 0 & $1\mathrm{e}{-4}$ & 4000 \\
Formula & sph\_harm11 & KAN-g5 & $5\mathrm{e}{-5}$ & 50 & $5\mathrm{e}{-5}$ & 4000 \\
Formula & sph\_harm02 & KAN-g5 & $2\mathrm{e}{-4}$ & 200 & $2\mathrm{e}{-5}$ & 4000 \\
Formula & sph\_harm12 & KAN-g5 & $2\mathrm{e}{-4}$ & 100 & $1\mathrm{e}{-4}$ & 4000 \\
Formula & sph\_harm22 & KAN-g5 & $2\mathrm{e}{-4}$ & 200 & $2\mathrm{e}{-5}$ & 4000 \\
\bottomrule
\end{tabular}
\end{table}

For the direct-training baseline, we select the optimizer learning rate and the number of training epochs $(\eta_{\mathrm{train}}, E_{\mathrm{train}})$ via grid search.
\Cref{tab:baseline-sweep-grid} summarizes the grids used, and \cref{tab:baseline-sweep-best} reports the best configuration selected for each experiment.

\begin{table}[t]
\centering
\caption{Direct-training hyperparameter sweep grids (universality experiment). Table entries list the sweep-grid values for $\eta_{\mathrm{train}}$ and $E_{\mathrm{train}}$.}
\label{tab:baseline-sweep-grid}
\scriptsize
\begin{tabular}{lllcc}
\toprule
Task & Dataset & Model & $\eta_{\mathrm{train}}$ & $E_{\mathrm{train}}$ \\
\midrule
Image & MNIST & MLP & $\{0.002,0.005,0.01\}$ & $\{50,100,200\}$ \\
Image & MNIST & CNN-8 & $\{0.005,0.01,0.02\}$ & $\{30,50,100\}$ \\
Image & CIFAR-10 & CNN-20 & $\{0.002,0.005,0.01\}$ & $\{100,200,400\}$ \\
\midrule
Graph & Cora & GCN & $\{0.002,0.005,0.01\}$ & $\{100,200,400\}$ \\
Graph & CiteSeer & GCN/GAT & $\{0.001,0.002,0.005\}$ & $\{100,200,400\}$ \\
Graph & PubMed & GCN & $\{0.002,0.005,0.01\}$ & $\{100,200,400\}$ \\
Graph & Cora & GAT & $\{0.005,0.01,0.02\}$ & $\{50,100,200\}$ \\
Graph & PubMed & GAT & $\{0.01,0.02,0.05\}$ & $\{100,200,400\}$ \\
\midrule
Text & AG News & Transformer-2L & $\{0.0005,0.001,0.002\}$ & $\{50,100,200\}$ \\
Text & IMDB & Transformer-1L & $\{0.0002,0.0005,0.001\}$ & $\{50,100,200\}$ \\
\midrule
Formula & ellipj/ellipkinc/ellipeinc & KAN-g5 & $\{0.01,0.02,0.05\}$ & $\{4000\}$ \\
Formula & jv/yv & KAN-g10 & $\{0.005,0.01,0.02\}$ & $\{4000\}$ \\
Formula & kv/iv & KAN-g5 & $\{0.005,0.01,0.02\}$ & $\{4000\}$ \\
Formula & lpmv0/lpmv1/lpmv2 & KAN-g5 & $\{0.02,0.05,0.1\}$ & $\{4000\}$ \\
Formula & sph\_harm01/sph\_harm02/sph\_harm11 & KAN-g5 & $\{0.02,0.05,0.1\}$ & $\{4000\}$ \\
Formula & sph\_harm12/sph\_harm22 & KAN-g5 & $\{0.02,0.05,0.1\}$ & $\{4000\}$ \\
\bottomrule
\end{tabular}
\end{table}

\begin{table}[t]
\centering
\caption{Direct-training best hyperparameters selected by grid search (universality experiment).}
\label{tab:baseline-sweep-best}
\scriptsize
\begin{tabular}{lllcc}
\toprule
Task & Dataset & Model & $\eta_{\mathrm{train}}$ & $E_{\mathrm{train}}$ \\
\midrule
Image & MNIST & MLP & 0.01 & 200 \\
Image & MNIST & CNN-8 & 0.005 & 50 \\
Image & CIFAR-10 & CNN-20 & 0.005 & 400 \\
\midrule
Graph & Cora & GCN & 0.01 & 100 \\
Graph & CiteSeer & GCN & 0.005 & 100 \\
Graph & PubMed & GCN & 0.005 & 200 \\
Graph & Cora & GAT & 0.02 & 100 \\
Graph & CiteSeer & GAT & 0.002 & 200 \\
Graph & PubMed & GAT & 0.02 & 200 \\
\midrule
Text & AG News & Transformer-2L & 0.0005 & 100 \\
Text & IMDB & Transformer-1L & 0.001 & 50 \\
\midrule
Formula & ellipj/ellipkinc/ellipeinc & KAN-g5 & 0.02 & 4000 \\
Formula & jv/yv & KAN-g10 & 0.02 & 4000 \\
Formula & kv/iv & KAN-g5 & 0.02 & 4000 \\
Formula & lpmv0 & KAN-g5 & 0.05 & 4000 \\
Formula & lpmv1/lpmv2 & KAN-g5 & 0.02 & 4000 \\
Formula & sph\_harm01/sph\_harm02/sph\_harm11 & KAN-g5 & 0.1 & 4000 \\
Formula & sph\_harm12/sph\_harm22 & KAN-g5 & 0.05 & 4000 \\
\bottomrule
\end{tabular}
\end{table}

\subsubsection{Scalability experiment}
For UHN and direct training, we reuse the hyperparameters selected on CNN-20 in the universality experiment (\Cref{tab:sweep-best,tab:baseline-sweep-best}) and apply them to CNN-32/44/56 without additional re-sweeps.

For HA \cite{ha2016hypernetworks}, we follow the original setting but use embedding size $188$ for a fairer capacity comparison.
We sweep the training learning rate and the number of training epochs $(\eta_{\mathrm{train}}, E_{\mathrm{train}})$ on CNN-20 only and reuse the selected setting for CNN-32/44/56.

For Chunked \cite{von2019continual}, we use hidden dimension $100$, embedding dimension $64$, and $2$ hidden layers.
We follow the same sweep protocol as HA.

\Cref{tab:scalability-sweep} reports the sweep grids and selected hyperparameters for HA and Chunked.

We use PyTorch default initialization \cite{paszke2019pytorch} for HA and Chunked, as it converged faster and yielded better results in our implementation.

\begin{table}[t]
    \centering
    \scriptsize
    \begin{tabular}{lcccc}
        \toprule
        Hypernetwork & $\eta_{\mathrm{train}}$ grid & $E_{\mathrm{train}}$ grid & Best $\eta_{\mathrm{train}}$ & Best $E_{\mathrm{train}}$ \\
        \midrule
        HA & $\{0.002,0.005,0.01\}$ & $\{200,400,800\}$ & 0.002 & 800 \\
        Chunked & $\{0.001,0.002,0.005\}$ & $\{200,400,800\}$ & 0.002 & 800 \\
        \bottomrule
    \end{tabular}
    \caption{Sweep grids and best hyperparameters for the hypernetwork baselines (scalability experiment).}
    \label{tab:scalability-sweep}
\end{table}

\section{Multi-Model Experiments}
\label{details-exp-multi-model}
This section provides additional experimental details supporting the multi-model experiments in \cref{sec:multi-model-experiment}, including model-set sampling details, model-set construction and splitting, hyperparameter selection, and supplementary scatter plots.

\subsection{Model Sampling Details}
\label{model-sampling-details}
For each multi-model experiment, we construct a finite model set $M$ by sampling architectures from the corresponding family under simple constraints (e.g., divisibility for GroupNorm \cite{wu2018group}).
In total, we construct four model sets: three CNN model sets for Exp-1--3 and one Transformer \cite{vaswani2017attention} model set for Exp-4.

\subsubsection{CNN family (Exp-1--3; CIFAR-10)}
We use a 4-stage CNN with stages indexed by $s\in\{0,1,2,3\}$. Let $K_s$ denote the number of convolutional layers in stage $s$ and $c_s$ the stage width (channels); under shortcuts \cite{he2016identity} (our default), $c_s$ is the (constant) number of output channels for all convolution layers within stage $s$.
Throughout this subsection, $\mathrm{Unif}\{a,\dots,b\}$ denotes the discrete uniform distribution over integers $a,a+1,\dots,b$.
We consider the following experiment-specific sampling schemes:
\begin{itemize}
    \item \textbf{Mixed depth (Exp-1).} Sample $K_1,K_2,K_3\sim \mathrm{Unif}\{6,\dots,10\}$ independently, with $K_0=1$, and fix $(c_0,c_1,c_2,c_3)=(16,16,32,64)$. We sample $|M|=100$ models.
    \item \textbf{Mixed width (Exp-2).} Fix $(K_0,K_1,K_2,K_3)=(1,6,6,6)$ and sample $c_0\sim \mathrm{Unif}\{16,\dots,32\}$, then set $c_1=c_0$, set $c_2\sim \mathrm{Unif}\{32,\dots,64\}$, and set $c_3\sim \mathrm{Unif}\{64,\dots,128\}$. We sample $|M|=500$ models.
    \item \textbf{Mixed depth\,$\times$\,width (Exp-3).} Sample $K_1,K_2,K_3\sim \mathrm{Unif}\{6,\dots,8\}$ and sample $c_0\sim \mathrm{Unif}\{16,\dots,32\}$, then set $c_1=c_0$, with $c_2\sim \mathrm{Unif}\{32,\dots,64\}$ and $c_3\sim \mathrm{Unif}\{64,\dots,128\}$. We sample $|M|=1000$ models.
\end{itemize}

\paragraph{Implementation details.}
Unless otherwise stated, all CNNs use $3\times 3$ convolutions and enable bias in both convolution and linear layers.
We use LeakyReLU with slope $0.1$ and GroupNorm with $g=4$ groups for all convolution layers except the first convolution layer.
We use no dropout \cite{srivastava2014dropout} in either convolution or linear layers.
Since GroupNorm with $g=4$ groups is used, we require $c_s\equiv 0\pmod g$. Residual shortcuts are enabled for all eligible layers except the output linear layer and the first convolution layer of stages $s\in\{0,2,3\}$ (so the first layer of stage $1$ uses a shortcut and we enforce $c_0=c_1$).
We only apply stage-wise downsampling at the first layer of stages $2$ and $3$, implemented via convolutional pooling (stride $2$).
The classifier head consists of adaptive average pooling followed by a single linear output layer (no activation, no normalization).

\subsubsection{Transformer family (Exp-4; AG News)}
In Exp-4, we sample a Transformer encoder stack by drawing the number of encoders $J\sim \mathrm{Unif}\{1,\dots,4\}$ and the number of heads $h\sim \mathrm{Unif}\{1,\dots,8\}$.
We sample the embedding dimension $d\sim \mathrm{Unif}\{32,\dots,128\}$ subject to $d\equiv 0\pmod h$, and for each encoder block $j\in\{1,\dots,J\}$ we sample the feed-forward depth $K_{\mathrm{j}}\sim \mathrm{Unif}\{1,\dots,3\}$.
We sample $|M|=1000$ models.

\paragraph{Implementation details.}
We use vocabulary size $V=5000$ and maximum sequence length $T_{\max}=128$.
All MHA layers and all linear layers use bias.
We apply dropout with rate $0.2$ in the embedding layer and in all MHA and feed-forward layers.
Residual shortcuts are enabled for all MHA and feed-forward layers. The final output linear layer uses neither dropout nor a shortcut.
Except for the first MHA layer and the final output linear layer, we apply LeakyReLU with slope $0.1$ and LayerNorm \cite{ba2016layer}; otherwise we use no activation and no normalization.
The final classifier head extracts the first token representation and applies a linear output layer.
Because shortcuts are used, the output dimension of all MHA and intermediate linear layers equals the embedding dimension $d$.

\subsection{Model-set construction and splitting}
\label{app:exp-modelset}

For each multi-model experiment, we sample a finite model set $M$ once from its model family detailed in \cref{model-sampling-details}. $M$ is further split into 2 subsets $M_{\mathrm{train}}$ and $M_{\mathrm{test}}$ for training and held-out evaluation. During hyperparameter selection, we reserve a validation subset $M_{\mathrm{val}} \subset M_{\mathrm{train}}$.

We also measure the hold-in performance. However, since evaluating every model in $M_{\mathrm{train}}$ is time consuming, we only use a hold-in subset $M'_{\mathrm{train}} \subset M_{\mathrm{train}}$ for evaluation, where $M'_{\mathrm{train}} \cap M_{\mathrm{val}} = \emptyset$.

\Cref{tab:multi-model-modelset} summarizes the split sizes for all 4 multi-model experiments.

\begin{table}[t]
    \centering
    \scriptsize
    \begin{tabular}{lrrrrr}
        \toprule
        Model family & $|M|$ & $|M_{\mathrm{train}}|$ & $|M_{\mathrm{test}}|$ & $|M_{\mathrm{val}}|$ & $|M'_{\mathrm{train}}|$ \\
        \midrule
        CNN Mixed Depth & 100 & 80 & 20 & 20 & 20 \\
        CNN Mixed Width & 500 & 450 & 50 & 50 & 50 \\
        CNN Mixed Depth\,$\times$\,Width & 1000 & 950 & 50 & 50 & 50 \\
        Transformer Mixed & 1000 & 950 & 50 & 50 & 50 \\
        \bottomrule
    \end{tabular}
    \caption{Model-set splits sizes for the multi-model experiments.}
    \label{tab:multi-model-modelset}
\end{table}

\subsection{Hyperparameter selection}
For each experiment, we sweep (i) the initialization-stage learning rate and steps $(\eta_{\mathrm{init}}, S_{\mathrm{init}})$ and (ii) the training-stage learning rate and epochs $(\eta_{\mathrm{train}}, E_{\mathrm{train}})$, following the sweep protocol described in \cref{details-method-sweep}.
\Cref{tab:multi-model-sweep-grid} lists the sweep grids, and \cref{tab:multi-model-sweep-best} lists the selected values.

\begin{table}[t]
    \centering
    \scriptsize
    \begin{tabular}{lccccc}
        \toprule
        Model family & $E_0$ & $\eta_{\mathrm{init}}$ & $S_{\mathrm{init}}$ & $\eta_{\mathrm{train}}$ & $E_{\mathrm{train}}$ \\
        \midrule
        CNN Mixed Depth & 200 & $\{5\mathrm{e}{-5},1\mathrm{e}{-4},2\mathrm{e}{-4}\}$ & $\{3200,6400,12800\}$ & $\{2\mathrm{e}{-5},5\mathrm{e}{-5},1\mathrm{e}{-4}\}$ & $\{800,1600,3200\}$ \\
        CNN Mixed Width & 200 & $\{5\mathrm{e}{-5},1\mathrm{e}{-4},2\mathrm{e}{-4}\}$ & $\{3200,6400,12800\}$ & $\{2\mathrm{e}{-5},5\mathrm{e}{-5},1\mathrm{e}{-4}\}$ & $\{800,1600,3200\}$ \\
        CNN Mixed Depth\,$\times$\,Width & 200 & $\{5\mathrm{e}{-5},1\mathrm{e}{-4},2\mathrm{e}{-4}\}$ & $\{3200,6400,12800\}$ & $\{2\mathrm{e}{-5},5\mathrm{e}{-5},1\mathrm{e}{-4}\}$ & $\{800,1600,3200\}$ \\
        Transformer Mixed & 50 & $\{5\mathrm{e}{-5},1\mathrm{e}{-4},2\mathrm{e}{-4}\}$ & $\{2000,4000,8000\}$ & $\{2\mathrm{e}{-5},5\mathrm{e}{-5},1\mathrm{e}{-4}\}$ & $\{50,100,200\}$ \\
        \bottomrule
    \end{tabular}
    \caption{Hyperparameter sweep grids for the multi-model experiments. Table entries list the sweep-grid values for $\eta_{\mathrm{init}}$, $S_{\mathrm{init}}$, $\eta_{\mathrm{train}}$, and $E_{\mathrm{train}}$.}
    \label{tab:multi-model-sweep-grid}
\end{table}

\begin{table}[t]
    \centering
    \scriptsize
    \begin{tabular}{lcccc}
        \toprule
        Model family & $\eta_{\mathrm{init}}$ & $S_{\mathrm{init}}$ & $\eta_{\mathrm{train}}$ & $E_{\mathrm{train}}$ \\
        \midrule
        CNN Mixed Depth & $2\mathrm{e}{-4}$ & 12800 & $5\mathrm{e}{-5}$ & 3200 \\
        CNN Mixed Width & $1\mathrm{e}{-4}$ & 12800 & $1\mathrm{e}{-4}$ & 3200 \\
        CNN Mixed Depth\,$\times$\,Width & $2\mathrm{e}{-4}$ & 3200 & $1\mathrm{e}{-4}$ & 3200 \\
        Transformer Mixed & $2\mathrm{e}{-4}$ & 8000 & $5\mathrm{e}{-5}$ & 100 \\
        \bottomrule
    \end{tabular}
    \caption{Best hyperparameters selected by grid search for the multi-model experiments.}
    \label{tab:multi-model-sweep-best}
\end{table}

\subsection{Multi-model scatter plots}
\begin{figure}[t]
  \centering
  \begin{subfigure}{0.48\linewidth}
    \centering
    \includegraphics[width=\linewidth]{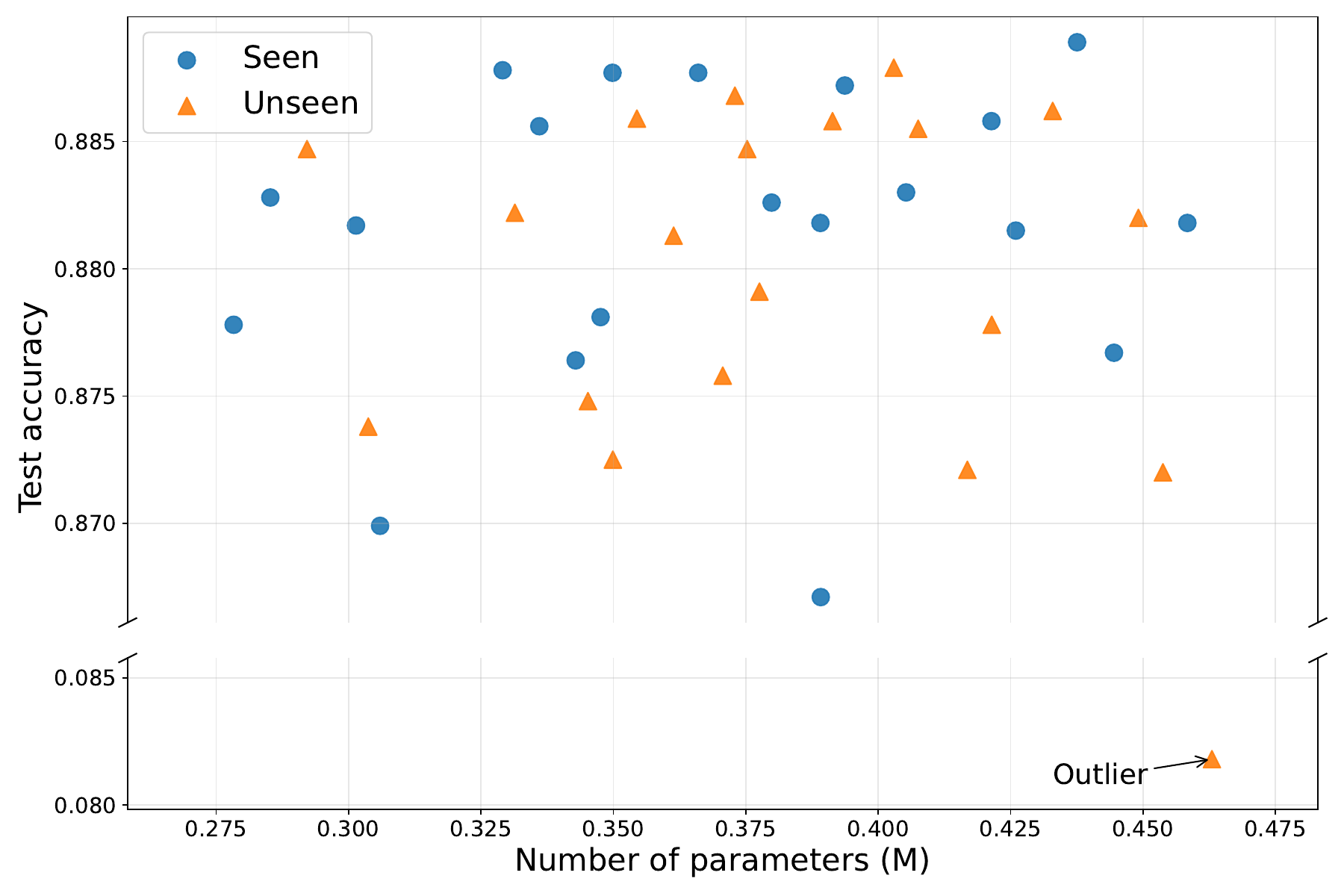}
    \caption{CNN Mixed Depth}
    \label{fig:scatter-cnn-depth}
  \end{subfigure}
  \hfill
  \begin{subfigure}{0.48\linewidth}
    \centering
    \includegraphics[width=\linewidth]{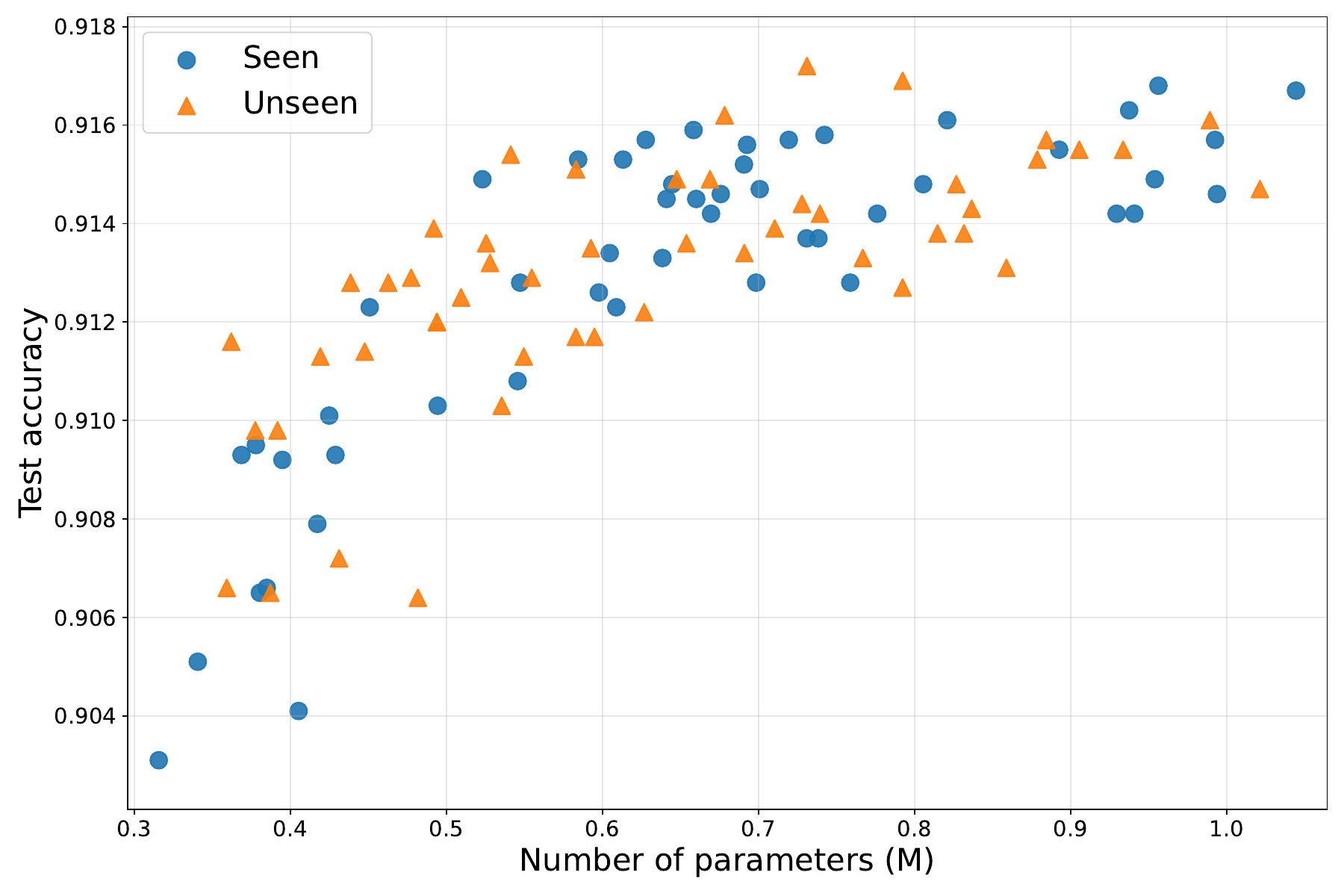}
    \caption{CNN Mixed Width}
    \label{fig:scatter-cnn-width}
  \end{subfigure}

  \vspace{0.4em}

  \begin{subfigure}{0.48\linewidth}
    \centering
    \includegraphics[width=\linewidth]{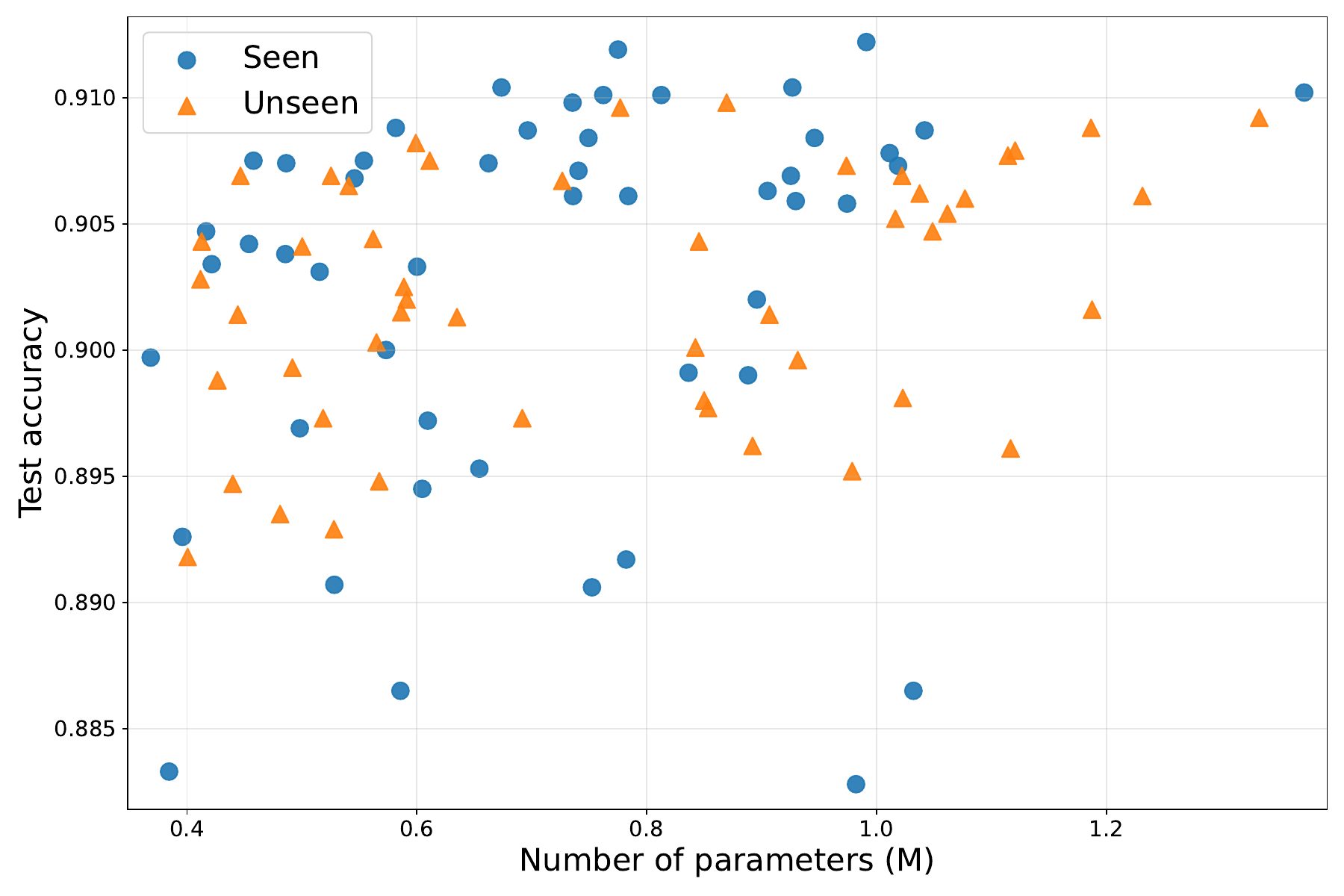}
    \caption{CNN Mixed Depth $\times$ Width}
    \label{fig:scatter-cnn-mixed_depth_and_width}
  \end{subfigure}
  \hfill
  \begin{subfigure}{0.48\linewidth}
    \centering
    \includegraphics[width=\linewidth]{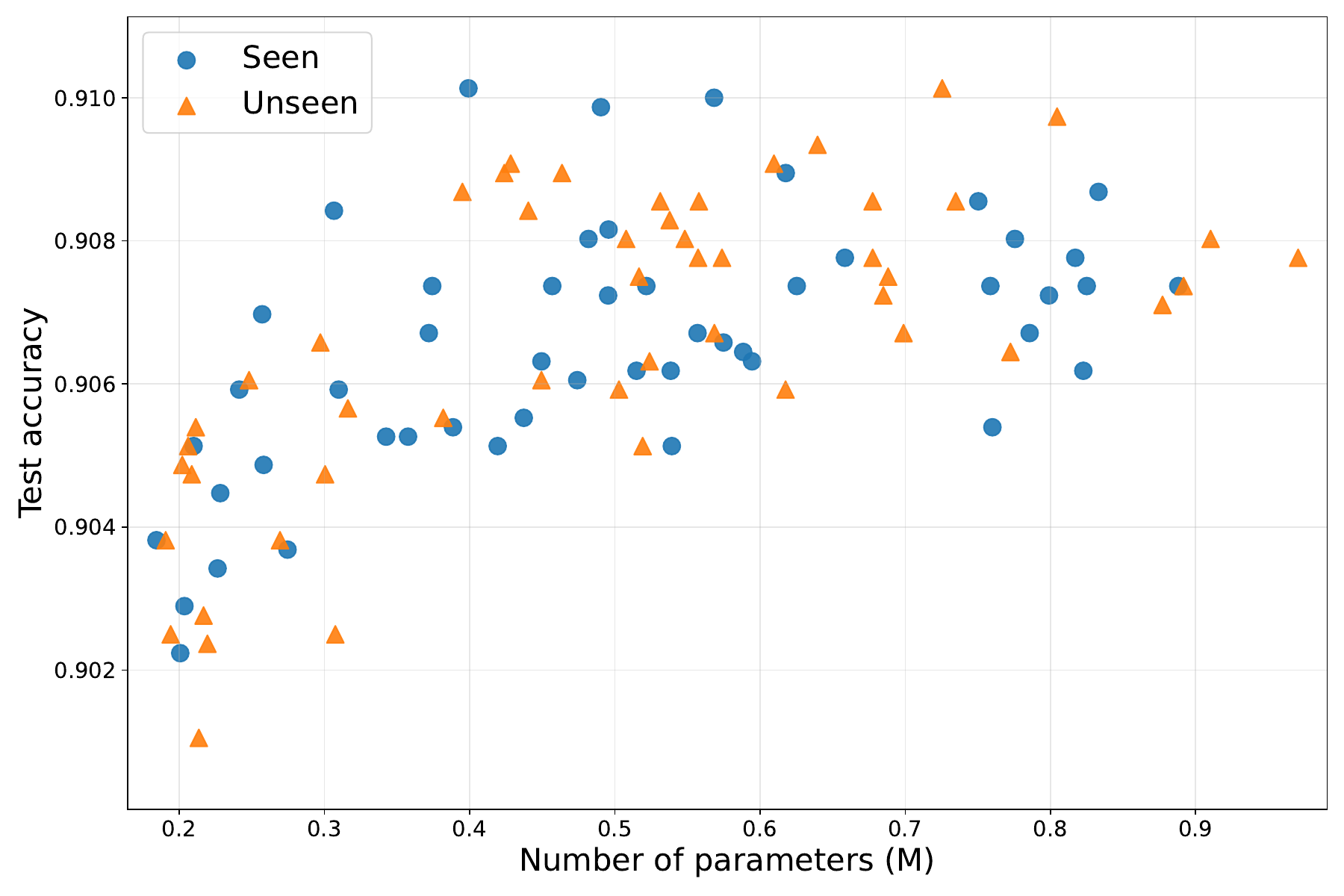}
    \caption{Transformer Mixed}
    \label{fig:scatter-transformer-mixed}
  \end{subfigure}
  \caption{Model-family performance scatter plots.}
  \label{fig:model-family-scatter}
\end{figure}

To illustrate the generalization ability of UHN to held-out architectures, we show the scatter plots (\cref{fig:model-family-scatter}) of the test performance of the models in hold-in set $M'_{\mathrm{train}}$ (``seen'') and held-out set $M_{\mathrm{test}}$ (``unseen'') generated by UHN versus the number of trainable parameters of the models.

\section{Multi-Task Experiment}
\label{app:details-exp-multi-task}
This section provides additional details on hyperparameter selection for the multi-task experiment in \cref{details-exp-multi-task}.

\subsection{Hyperparameter selection.}
Our goal is to select a single multi-task hyperparameter configuration that works well across all tasks.
To avoid solutions that perform well on average but fail on an individual task, we enforce task-specific validation guardrails (\cref{tab:multitask-guardrails}) and discard any configuration that violates \emph{any} task's guardrail.
For metrics where higher is better (e.g., classification accuracy), we require the validation metric to be at least the guardrail; for metrics where lower is better (e.g., RMSE), we require it to be at most the guardrail.

We sweep (i) the initialization-stage learning rate and steps $(\eta_{\mathrm{init}}, S_{\mathrm{init}})$ and (ii) the training-stage learning rate and steps $(\eta_{\mathrm{train}}, S_{\mathrm{train}})$, following the sweep protocol described in \cref{details-method-sweep}. \cref{tab:multitask-sweep} summarizes the sweep grid and the selected hyperparameters.

\begin{table}[t]
    \centering
    \scriptsize
    \begin{tabular}{lllc}
        \toprule
        Task & Dataset & Model & Guardrail (val) \\
        \midrule
        Image & MNIST & MLP & 0.95 \\
        Image & CIFAR-10 & CNN-44 & 0.85 \\
        Graph & Cora & GCN & 0.75 \\
        Graph & PubMed & GAT & 0.75 \\
        Text & AG News & Transformer-2L & 0.85 \\
        Formula & kv & KAN-g5 & $5\mathrm{e}{-2}$ \\
        \bottomrule
    \end{tabular}
    \caption{Task-specific validation guardrails used during hyperparameter selection in the multi-task setting.}
    \label{tab:multitask-guardrails}
\end{table}

\begin{table}[t]
    \centering
    \scriptsize
    \begin{tabular}{lcccc}
        \toprule
         & $\eta_{\mathrm{init}}$ & $S_{\mathrm{init}}$ & $\eta_{\mathrm{train}}$ & $S_{\mathrm{train}}$ \\
        \midrule
        Grid ($S_0=40000$) & $\{5\mathrm{e}{-5},1\mathrm{e}{-4},2\mathrm{e}{-4}\}$ & $\{500,1000,2000\}$ & $\{1\mathrm{e}{-5},2\mathrm{e}{-5},5\mathrm{e}{-5}\}$ & $\{100000,200000,300000\}$ \\
        Selected & $1\mathrm{e}{-4}$ & 500 & $2\mathrm{e}{-5}$ & 200000 \\
        \bottomrule
    \end{tabular}
    \caption{Hyperparameter sweep grid and selected hyperparameters for multi-task training (subject to the guardrails in \cref{tab:multitask-guardrails}).}
    \label{tab:multitask-sweep}
\end{table}

\section{Recursive Experiments}
\label{app:recursive-details}
This section provides implementation details for the recursive experiments in \cref{details-exp-recursive-task}, including (i) the architecture of generated UHNs $H_i$ ($i\ge 1$), (ii) additional training details, and (iii) hyperparameter selection.

\subsection{Generated UHN architecture}
Each generated hypernetwork $H_i$ ($i\ge 1$) follows the architectural template in \cref{recursive-hypernetwork-model} with the following settings.

\paragraph{Two-branch parameter generator}
Each generated UHN consists of two branches whose outputs are summed before the final readout.

\paragraph{Index branch}
The index branch takes Gaussian Fourier features \cite{tancik2020fourier} of the normalized index descriptor with $F_{\mathbf{v}}=1024$ frequencies.
It applies an input linear layer followed by 4 linear layers with hidden dimension $d=64$, each with LeakyReLU (slope $0.1$), LayerNorm \cite{ba2016layer}, and a shortcut connection \cite{he2016identity}.

\paragraph{Task-Structure branch}
The task-structure branch takes Gaussian Fourier features of the normalized task-structure descriptors with $F_{\mathbf{u}}=32$ frequencies.
It applies a multi-head attention \cite{vaswani2017attention} layer with $h=4$ heads (with a shortcut connection), followed by two linear layers with LeakyReLU, LayerNorm, and shortcut connections.
The resulting sequence is average-pooled across the sequence dimension, followed by a linear layer with output size 64 and then a second linear layer with LeakyReLU.

\paragraph{Readout}
We sum the index branch and task-structure branch features and apply a final linear layer with LeakyReLU to predict each target scalar weight.

We share descriptor normalization statistics and Fourier matrices across recursion levels (i.e., all $H_i$ use the same ones as the root $H_0$).

\subsection{Additional training details}
Since deeper recursion increases training instability, we apply gradient clipping \cite{pascanu2013difficulty} with maximum norm $0.01$ in \emph{both} the initialization and training phases for all recursion depths (including $K=1$).
During recursive initialization, we reset the optimizer and learning-rate scheduler at each initialization level.

\subsection{Hyperparameter selection}
We sweep (i) the initialization-stage learning rate and steps $(\eta_{\mathrm{init}}, S_{\mathrm{init}})$ and (ii) the training-stage learning rate and steps $(\eta_{\mathrm{train}}, S_{\mathrm{train}})$ for recursion depth $K=1$, following the sweep protocol described in \cref{details-method-sweep}. We set the number of initialization steps allocated to each recursion level as $S_{\mathrm{lvl}} = S_{\mathrm{init}}/2$, and reuse the same $S_{\mathrm{lvl}}$ for all recursion depths.
\Cref{tab:recursive-sweep} summarizes the sweep grid and the selected best configuration.

\begin{table}[t]
    \centering
    \scriptsize
    \begin{tabular}{lcccc}
        \toprule
         & $\eta_{\mathrm{init}}$ & $S_{\mathrm{init}}$ & $\eta_{\mathrm{train}}$ & $S_{\mathrm{train}}$ \\
        \midrule
        Grid ($S_0=30000$) & $\{5\mathrm{e}{-5},1\mathrm{e}{-4},2\mathrm{e}{-4}\}$ & $\{1000,2000,4000\}$ & $\{2\mathrm{e}{-5},5\mathrm{e}{-5},1\mathrm{e}{-4}\}$ & $\{15000,30000,60000\}$ \\
        Selected & $1\mathrm{e}{-4}$ & 4000 & $2\mathrm{e}{-5}$ & 30000 \\
        \bottomrule
    \end{tabular}
    \caption{Hyperparameter sweep grid and selected hyperparameters for the recursive setting with recursion depth $K=1$.}
    \label{tab:recursive-sweep}
\end{table}

Let $S^{(K=1)}_{\mathrm{init}}$, $S^{(K=1)}_{\mathrm{train}}$, $\eta^{(K=1)}_{\mathrm{init}}$, and $\eta^{(K=1)}_{\mathrm{train}}$ denote the selected step budgets and learning rates at $K=1$.
For depths $K>1$, we set the total initialization steps to $S_{\mathrm{init}} = (K+1)\,S^{(K=1)}_{\mathrm{init}}/2$, reuse $S_{\mathrm{train}} = S^{(K=1)}_{\mathrm{train}}$, and reduce learning rates to maintain stability.
During initialization, we use a per-level learning rate schedule
\begin{equation}
\eta^{(k)}_{\mathrm{init}} =
\begin{cases}
\eta^{(K=1)}_{\mathrm{init}}/5, & k\in\{0,1\},\\
\eta^{(K=1)}_{\mathrm{init}}/10, & k=2,\\
\eta^{(K=1)}_{\mathrm{init}}/40, & k=3,
\end{cases}
\end{equation}
where $k\in\{0,\dots,K\}$ indexes the recursion level, and $\eta^{(k)}_{\mathrm{init}}$ denotes learning rate for level-$k$ initialization.
We use $1000$ warmup steps at each initialization level.
During training, we use a reduced training-stage learning rate for deeper recursion: $\eta^{(K=1)}_{\mathrm{train}}/2$ for $K=2$ and $\eta^{(K=1)}_{\mathrm{train}}/8$ for $K=3$.

%%%%%%%%%%%%%%%%%%%%%%%%%%%%%%%%
% Appendix: Ablations
%%%%%%%%%%%%%%%%%%%%%%%%%%%%%%%%
\section{Ablation Experiment}
\label{app:ablations}
We ablate key design choices of UHN, including the task-structure encoder, initialization, index encoding, and hypernetwork capacity.

\subsection{Index Encoding}
\subsubsection{Setup}
UHN generates scalar parameters by conditioning on index descriptors.
Since MLPs exhibit a spectral bias towards low-frequency functions \cite{tancik2020fourier}, the choice of index encoding can be critical when the target parameter field varies rapidly across indices.
Note that the index encoding is applied to the normalized index descriptor $\hat{\mathbf{v}}_i\in\mathbb{R}^{10}$ (computed as in \cref{details-method-normal}).
We test the effect of different index encoding schemes by applying an encoding $\gamma(\cdot)$ to $\hat{\mathbf{v}}_i$:
\begin{enumerate}
    \item \textbf{Raw encoding (no mapping).} $\gamma_{\mathrm{raw}}(\hat{\mathbf{v}}_i)=\hat{\mathbf{v}}_i$.
    \item \textbf{Positional encoding \cite{mildenhall2021nerf}.}
    \[
    \gamma_{\mathrm{pos}}(\hat{\mathbf{v}}_i) = \big[\cos(\omega^0 \hat{\mathbf{v}}_i)^{\top},\; \sin(\omega^0 \hat{\mathbf{v}}_i)^{\top},\; \dots,\; \cos(\omega^{n-1} \hat{\mathbf{v}}_i)^{\top},\; \sin(\omega^{n-1} \hat{\mathbf{v}}_i)^{\top}\big]^{\top},
    \]
    where $\omega^j = \sigma^{j/n}$ for $j\in\{0,\dots,n-1\}$; we set the number of frequencies $n=32$, and scale $\sigma=100$.
    \item \textbf{Gaussian Fourier features (GFF) \cite{tancik2020fourier}.}
    $\gamma_{\mathbf{B}_{\mathbf{v}}}(\hat{\mathbf{v}}_i) = \big[\cos(\mathbf{B}_{\mathbf{v}}\hat{\mathbf{v}}_i)^{\top},\; \sin(\mathbf{B}_{\mathbf{v}}\hat{\mathbf{v}}_i)^{\top}\big]^{\top}$,
    where $\mathbf{B}_{\mathbf{v}}\in\mathbb{R}^{F_{\mathbf{v}}\times 10}$ with every entry sampled from $\mathcal{N}(0,\sigma^2)$; we set the number of Fourier-feature frequencies $F_{\mathbf{v}}=2048$ and scale $\sigma=100$.
\end{enumerate}

We run the ablation using the single-model experiment setting (\cref{single-model-exp}) on CIFAR-10 where the base model is CNN-20, and only change the encoding scheme.
Gaussian Fourier features are the default index encoding scheme in our single-model experiment; its result in \Cref{tab:enc-abl-perf} therefore corresponds to the single-model experiment setting.
For Raw and Positional encodings, we re-sweep the training learning rate $\eta_{\mathrm{train}}$ over $\{5\mathrm{e}{-5},1\mathrm{e}{-4},2\mathrm{e}{-4}\}$ and select $2\mathrm{e}{-4}$.
Since Raw and Positional encodings are less stable in our experiments, we apply gradient clipping \cite{pascanu2013difficulty} with maximum norm $1.0$ during the training phase for these two settings (no clipping during the initialization phase).

\subsubsection{Results.}
\Cref{tab:enc-abl-perf} reports test accuracy (mean $\pm$ std over $n=3$ seeds) and the number of trainable parameters of the UHN.

\begin{table}[t]
    \centering
    \scriptsize
    \begin{tabular}{lrl}
        \toprule
        Encoding type & \#Params & Acc. \\
        \midrule
        Raw & 68{,}629 & $0.6642 \pm 0.0199$ \\
        Positional & 149{,}301 & $0.8677 \pm 0.0034$ \\
        GFF & 612{,}117 & $0.8993 \pm 0.0016$ \\
        \bottomrule
    \end{tabular}
    \caption{Index-encoding ablation results. \#Params denotes the number of trainable parameters of UHN; Acc. denotes test accuracy.}
    \label{tab:enc-abl-perf}
\end{table}

\subsubsection{Analysis}
\begin{figure*}[t]
  \centering
  \begin{subfigure}{0.48\textwidth}
    \centering
    \includegraphics[width=\linewidth]{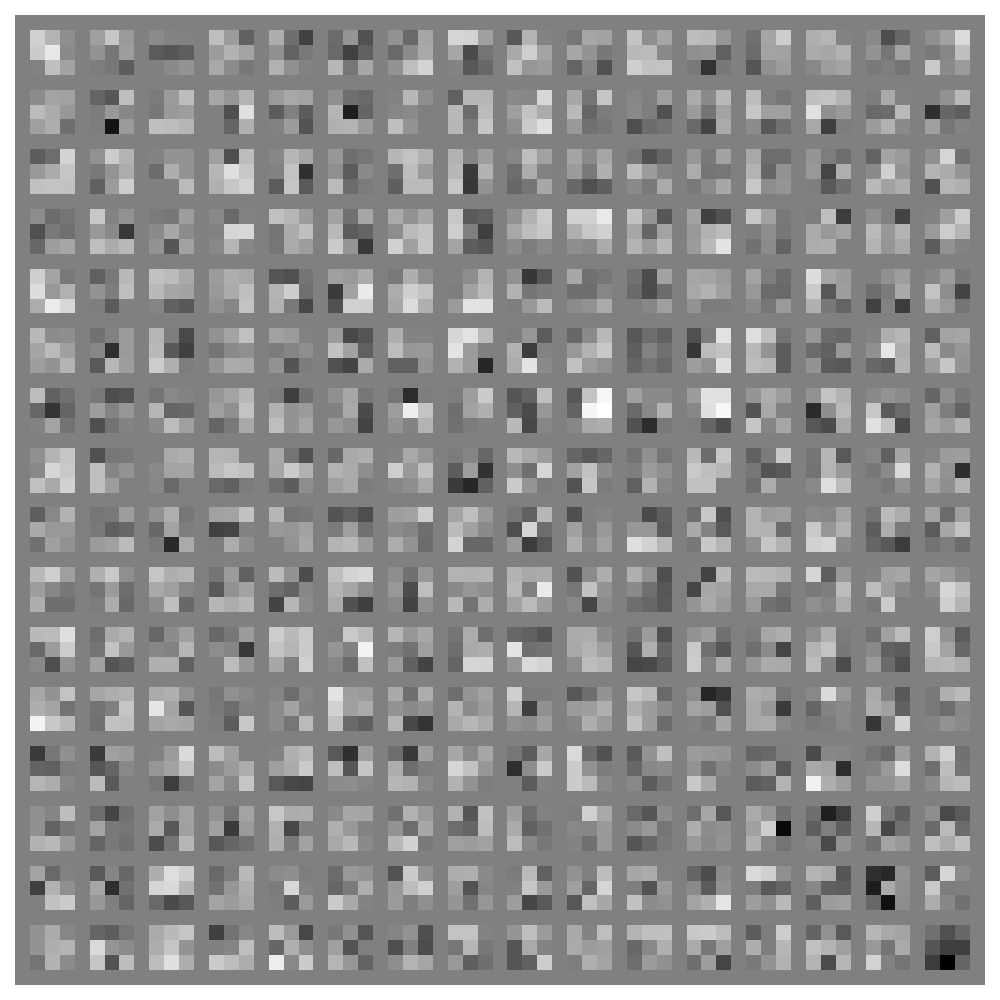}
    \caption{Gaussian Fourier features}
    \label{fig:kernel-vis-gff}
  \end{subfigure}
  \hfill
  \begin{subfigure}{0.48\textwidth}
    \centering
    \includegraphics[width=\linewidth]{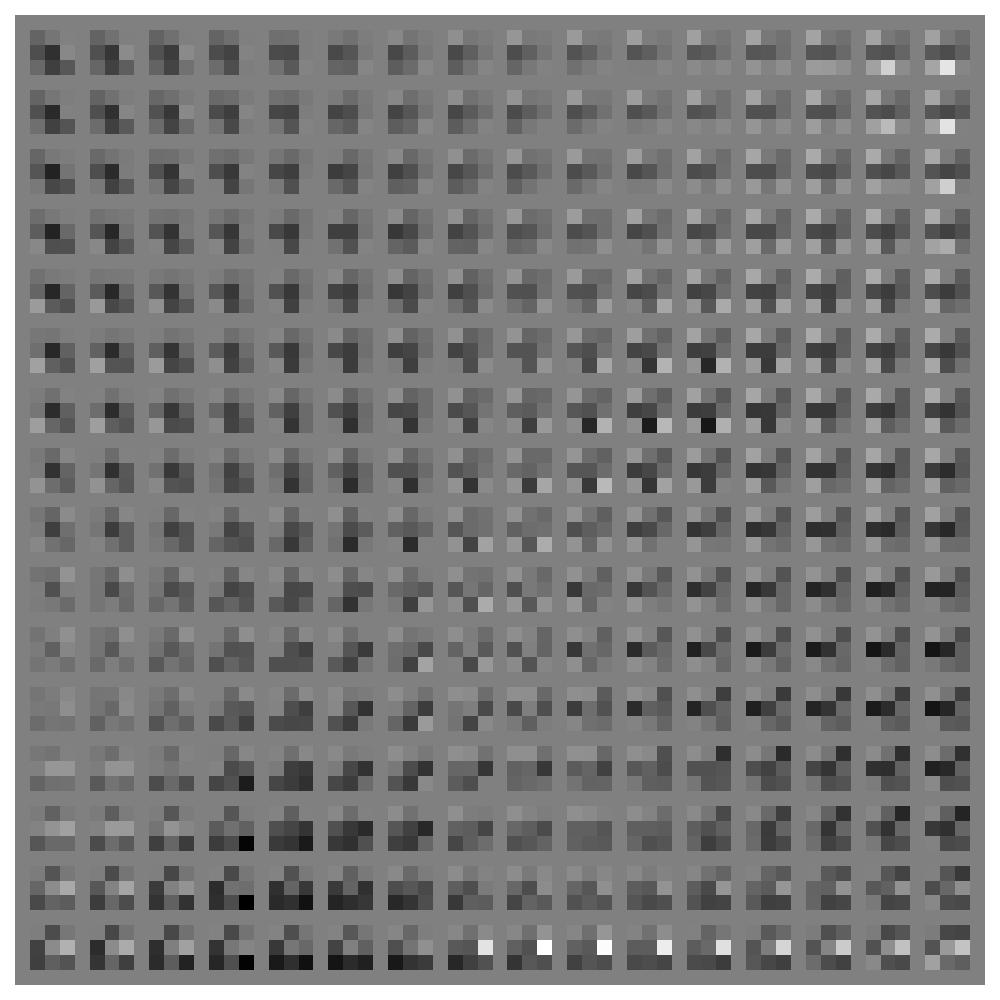}
    \caption{Raw encoding}
    \label{fig:kernel-vis-raw}
  \end{subfigure}
  \caption{Visualization of the third convolution-layer kernels of the generated base model.}
  \label{fig:kernel-vis}
\end{figure*}

Overall, more expressive encodings improve performance, at the cost of a larger UHN: Raw underperforms substantially, positional encoding provides a large gain, and Gaussian Fourier features achieve the best accuracy.
We also visualize the kernels of the third convolution layer of the generated base model (\cref{fig:kernel-vis}). We observe that Raw encoding produces low-variance (less diverse) kernels, whereas Gaussian Fourier features yields more structured and diverse kernels.
This is consistent with the intuition that Gaussian Fourier features expose higher-frequency basis functions \cite{tancik2020fourier} to the hypernetwork, mitigating spectral bias and enabling it to model rapidly varying parameter fields over index space.

\subsection{UHN Capacity}
\subsubsection{Setup}
We study how UHN performance scales with generator capacity, testing whether accuracy improves predictably as we increase (i) the richness of the index encoding, (ii) the width of the generator, or (iii) its depth.
We run the UHN capacity ablation in the single-model setting (\cref{single-model-exp}) on CIFAR-10 where the base model is CNN-20.
Starting from the default configuration $(F_{\mathbf{v}}, d, N_{\mathrm{blk}})=(2048,128,2)$, we vary one of three capacity-related hyperparameters while keeping the other two fixed: the number of Fourier-feature \cite{tancik2020fourier} frequencies $F_{\mathbf{v}}$ for index encoding, the UHN hidden dimension $d$, and the number of UHN residual blocks $N_{\mathrm{blk}}$.

\subsubsection{Results}
For each configuration, we report test accuracy (mean $\pm$ std over $n=3$ seeds) and the number of trainable parameters of the UHN.
\Cref{tab:cap} summarizes the results.

\begin{table}[t]
    \centering
    \scriptsize
    \begin{tabular}{rrl rrl rrl}
        \toprule
        \multicolumn{3}{c}{$F_{\mathbf{v}}$} &
        \multicolumn{3}{c}{$d$} &
        \multicolumn{3}{c}{$N_{\mathrm{blk}}$} \\
        \cmidrule(lr){1-3} \cmidrule(lr){4-6} \cmidrule(lr){7-9}
        Value & \#Params & Acc. & Value & \#Params & Acc. & Value & \#Params & Acc. \\
        \midrule
        256 & 135{,}445 & $0.8894 \pm 0.0020$ & 32 & 156{,}117 & $0.8900 \pm 0.0037$ & 0 & 545{,}045 & $0.8901 \pm 0.0028$ \\
        512 & 203{,}541 & $0.8955 \pm 0.0005$ & 64 & 299{,}925 & $0.8960 \pm 0.0015$ & 1 & 578{,}581 & $0.8995 \pm 0.0008$ \\
        1024 & 339{,}733 & $0.8964 \pm 0.0012$ & \underline{128} & \underline{612{,}117} & \underline{$0.8993 \pm 0.0016$} & \underline{2} & \underline{612{,}117} & \underline{$0.8993 \pm 0.0016$} \\
        \underline{2048} & \underline{612{,}117} & \underline{$0.8993 \pm 0.0016$} & 256 & 1{,}334{,}805 & $0.9019 \pm 0.0041$ & 3 & 645{,}653 & $0.9009 \pm 0.0010$ \\
        4096 & 1{,}156{,}885 & $0.9018 \pm 0.0012$ & -- & -- & -- & -- & -- & -- \\
        \bottomrule
    \end{tabular}
    \caption{Capacity ablation results. We vary one hyperparameter at a time: the number of Fourier-feature frequencies $F_{\mathbf{v}}$ for index encoding (left), the UHN hidden dimension $d$ (middle), and the number of UHN residual blocks $N_{\mathrm{blk}}$ (right). Underlined entries indicate the single-model experiment setting (\cref{single-model-exp}). \#Params denotes the number of trainable parameters of UHN; Acc. denotes test accuracy.}
    \label{tab:cap}
\end{table}

\subsubsection{Analysis}
Increasing capacity generally improves performance, at the cost of more trainable parameters.
Increasing the number of Fourier-feature frequencies $F_{\mathbf{v}}$ for index encoding and the UHN hidden dimension $d$ yields consistent improvements, supporting the view that richer high-frequency index features and higher-dimensional hidden representations are critical for accurate weight generation.
By contrast, increasing the number of UHN residual blocks $N_{\mathrm{blk}}$ yields a noticeable improvement when going from $0$ to $1$ block, but only marginal improvements beyond one block, suggesting that moderate UHN depth is sufficient in this setting.
Overall, the results suggest that the default configuration $(F_{\mathbf{v}}, d, N_{\mathrm{blk}})=(2048,128,2)$ offers a strong accuracy--parameter trade-off among the tested settings.
This scaling behavior supports the view that UHN can be strengthened by increasing capacity along standard axes without changing the overall generation mechanism.

\subsection{Task-Structure Encoder}
\label{sec:structure-encoding}
\subsubsection{Setup}
UHN can condition on explicit descriptors of the target architecture and task via a \emph{task-structure encoder}. However, it is unclear how much this additional conditioning contributes beyond the index-based generator.
To isolate the task-structure encoder's effect, we disable it while keeping the remaining components and training protocol unchanged, and evaluate in two settings:
(i) \textbf{multi-model training} on two representative model families (CNN Mixed Width and Transformer Mixed) following \cref{sec:multi-model-experiment}; and
(ii) \textbf{multi-task training} across heterogeneous tasks following \cref{details-exp-multi-task}.

\subsubsection{Results}
For multi-model training, we report test accuracy averaged over architectures in $M'_{\mathrm{train}}$ (the subset of training architectures used for hold-in evaluation; ``seen'') and $M_{\mathrm{test}}$ (held-out test architectures; ``unseen'').
\Cref{tab:struct-enc-multi-model} reports the mean and standard deviation of these averages over 3 random seeds.

For multi-task training, we use UHN to generate each task's base model and evaluate test performance on that task.
\Cref{tab:struct-enc-multi-task} reports the mean and standard deviation over 3 random seeds.
For classification tasks, we report \emph{test accuracy} (higher is better); for formula regression, we report \emph{test RMSE} (lower is better).

\begin{table}[t]
    \centering
    \scriptsize
    \begin{tabular}{llrrr}
        \toprule
        Model family & TSE & \#Params & Seen Acc. & Unseen Acc. \\
        \midrule
        CNN Mixed Width & w/ & 663{,}151 & $0.9145 \pm 0.0014$ & $0.9145 \pm 0.0013$ \\
        CNN Mixed Width & w/o & 612{,}117 & $0.9143 \pm 0.0012$ & $0.9141 \pm 0.0013$ \\
        \midrule
        Transformer Mixed & w/ & 663{,}151 & $0.9063 \pm 0.0004$ & $0.9066 \pm 0.0002$ \\
        Transformer Mixed & w/o & 612{,}117 & $0.9073 \pm 0.0021$ & $0.9074 \pm 0.0023$ \\
        \bottomrule
    \end{tabular}
    \caption{Task-structure-encoder (TSE) ablation in the multi-model setting. ``w/'' and ``w/o'' indicate whether the task-structure encoder is enabled or disabled. \#Params is the number of trainable parameters of UHN. ``Seen Acc.'' averages test accuracy over $M'_{\mathrm{train}}$ (the subset of training architectures used for hold-in evaluation) and ``Unseen Acc.'' averages test accuracy over $M_{\mathrm{test}}$ (held-out test architectures).}
    \label{tab:struct-enc-multi-model}
\end{table}

\begin{table}[t]
    \centering
    \scriptsize
    \begin{tabular}{lll ll}
        \toprule
        Task & Dataset & Model & UHN perf. (multi-task, w/ TSE) & UHN perf. (multi-task, w/o TSE) \\
        \midrule
        Image & MNIST & MLP & $0.9786 \pm 0.0012$ & $0.9783 \pm 0.0011$ \\
        Image & CIFAR-10 & CNN-44 & $0.8927 \pm 0.0011$ & $0.8895 \pm 0.0015$ \\
        Graph & Cora & GCN & $0.7930 \pm 0.0053$ & $0.7927 \pm 0.0021$ \\
        Graph & PubMed & GAT & $0.7697 \pm 0.0031$ & $0.7703 \pm 0.0042$ \\
        Text & AG News & Transformer-2L & $0.9062 \pm 0.0005$ & $0.9057 \pm 0.0002$ \\
        \midrule
        Formula (RMSE) & kv & KAN-g5 & $0.0172 \pm 0.0112$ & $0.0082 \pm 0.0018$ \\
        \bottomrule
    \end{tabular}
    \caption{Task-structure-encoder (TSE) ablation in the multi-task setting. ``w/ TSE'' and ``w/o TSE'' indicate whether the task-structure encoder is enabled or disabled. ``Perf.'' denotes test accuracy for classification (higher is better) and test RMSE for formula regression (lower is better).}
    \label{tab:struct-enc-multi-task}
\end{table}

\subsubsection{Analysis}
In our experiments, enabling the task-structure encoder often stabilizes early-stage training progress. In terms of final test performance, the effect is small and task-dependent.
In multi-model training, enabling the task-structure encoder yields a slight gain for CNN Mixed Width, whereas for Transformer Mixed it leads to a slight degradation.
In multi-task training, the task-structure encoder improves performance on some tasks (e.g., CIFAR-10), has a negligible effect on others (e.g., the graph tasks), and degrades performance on formula regression (RMSE increases).
Overall, the task-structure encoder primarily improves \emph{optimization} (stabilizing early-stage training) while having a slight and sometimes mixed effect on final test performance. For multi-model settings, this may be because different models within a family are often highly similar, enabling effective hard parameter sharing even without additional structure information. For multi-task settings, in our setup only a small subset of parameters is hard-shared even without explicit structure or task descriptors, while the remaining parameters can be disambiguated by their indices; consequently, the generator can implicitly learn which parameters belong to which task during training, further weakening the marginal benefit of the task-structure encoder. These observations support our claim that UHN's core generality comes from the shared index-based generator, with the task-structure encoder acting as an auxiliary conditioning module.

\subsection{Initialization}
\subsubsection{Setup}
Mismatches between the initial weight distribution of the base model weights generated by UHN and standard neural network initializations (e.g., Glorot initialization \cite{glorot2010understanding}) can destabilize optimization.
We therefore include an \emph{initialization stage} that calibrates the generated weights to a target distribution before the main training begins (by default, we match PyTorch's standard parameter initialization \cite{paszke2019pytorch} for the corresponding layer unless otherwise specified).

We evaluate the effect of this stage by disabling it (setting the initialization-stage learning rate and steps $(\eta_{\mathrm{init}}, S_{\mathrm{init}})$ to $0$) while keeping all other settings unchanged.
We report results in three settings:
(i) \textbf{multi-model training} on two representative model families (CNN Mixed Width and Transformer Mixed) following \cref{sec:multi-model-experiment};
(ii) \textbf{multi-task training} across heterogeneous tasks following \cref{details-exp-multi-task}; and
(iii) \textbf{recursive training} with recursion depth $K=1$ following \cref{details-exp-recursive-task}.

\subsubsection{Results}
For multi-model training, we report test accuracy averaged over architectures in $M'_{\mathrm{train}}$ (the subset of training architectures used for hold-in evaluation; ``seen'') and $M_{\mathrm{test}}$ (held-out test architectures; ``unseen'').
\Cref{tab:init-multimodel} reports the mean and standard deviation of these averages over 3 random seeds.

For multi-task training, we use UHN to generate each task's base model and evaluate test performance on that task.
\Cref{tab:init-multitask} reports the mean and standard deviation over 3 random seeds.
For classification tasks, we report \emph{test accuracy} (higher is better); for formula regression, we report \emph{test RMSE} (lower is better).

For recursive training, we observe that disabling initialization causes training to diverge.

\begin{table}[t]
    \centering
    \scriptsize
    \begin{tabular}{llrr}
        \toprule
        Model family & init & Seen Acc. & Unseen Acc. \\
        \midrule
        CNN Mixed Width & w/ & $0.9145 \pm 0.0014$ & $0.9145 \pm 0.0013$ \\
        CNN Mixed Width & w/o & $0.9130 \pm 0.0011$ & $0.9130 \pm 0.0011$ \\
        \midrule
        Transformer Mixed & w/ & $0.9063 \pm 0.0004$ & $0.9066 \pm 0.0002$ \\
        Transformer Mixed & w/o & $0.8966 \pm 0.0095$ & $0.8963 \pm 0.0096$ \\
        \bottomrule
    \end{tabular}
    \caption{Initialization (init) ablation in the multi-model setting. ``w/'' and ``w/o'' indicate whether the initialization is enabled or disabled. ``Seen Acc.'' averages test accuracy over $M'_{\mathrm{train}}$ (the subset of training architectures used for hold-in evaluation) and ``Unseen Acc.'' averages test accuracy over $M_{\mathrm{test}}$ (held-out test architectures).}
    \label{tab:init-multimodel}
\end{table}

\begin{table}[t]
    \centering
    \scriptsize
    \begin{tabular}{lll ll}
        \toprule
        Task & Dataset & Model & UHN perf. (multi-task, w/ init) & UHN perf. (multi-task, w/o init) \\
        \midrule
        Image & MNIST & MLP & $0.9786 \pm 0.0012$ & $0.9780 \pm 0.0016$ \\
        Image & CIFAR-10 & CNN-44 & $0.8927 \pm 0.0011$ & $0.8875 \pm 0.0023$ \\
        Graph & Cora & GCN & $0.7930 \pm 0.0053$ & $0.7750 \pm 0.0040$ \\
        Graph & PubMed & GAT & $0.7697 \pm 0.0031$ & $0.7690 \pm 0.0017$ \\
        Text & AG News & Transformer-2L & $0.9062 \pm 0.0005$ & $0.9051 \pm 0.0025$ \\
        \midrule
        Formula (RMSE) & kv & KAN-g5 & $0.0172 \pm 0.0112$ & $0.0196 \pm 0.0085$ \\
        \bottomrule
    \end{tabular}
    \caption{Initialization (init) ablation in the multi-task setting. ``w/ init'' and ``w/o init'' indicate whether the initialization is enabled or disabled. ``Perf.'' denotes test accuracy for classification (higher is better) and test RMSE for formula regression (lower is better).}
    \label{tab:init-multitask}
\end{table}

\subsubsection{Analysis}
\begin{figure}[t]
    \centering
    \includegraphics[width=0.6\linewidth]{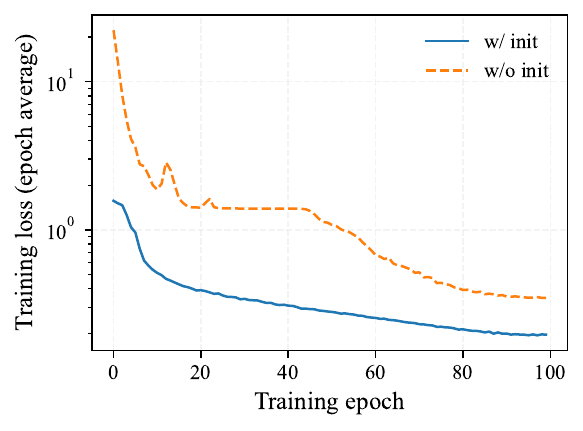}
    \caption{Multi-model Transformer Mixed epoch-averaged training loss with and without the initialization stage (``w/ init'' and ``w/o init''; log-scale $y$-axis). Loss is averaged over all training samples within each epoch.}
    \label{fig:init-multimodel-transformer-loss}
\end{figure}

In our experiments, initialization always improves optimization stability and accelerates early-stage convergence.
However, its effect on \emph{final} test performance is setting- and task-dependent.
In multi-model training, initialization consistently improves early-stage convergence and stability, and often improves final performance; \cref{fig:init-multimodel-transformer-loss} shows that initialization substantially reduces the epoch-averaged training loss in the early phase for Transformer Mixed.
In multi-task training, initialization improves performance across all tasks, though the magnitude of the gains varies by task.
In the recursive setting, initialization is essential---without it, generated downstream weights can have excessively large magnitude, leading to exploding activations and unstable gradients.
Overall, initialization speeds up early convergence across settings and is essential in the recursive case, and it often improves final performance.

\section{Extended Discussion}
\label{app:discussion}
This section expands the main-text discussion by providing additional empirical insights, limitations, and future directions.
\paragraph{Evidence of shared structure across models and tasks.}
Our multi-model and multi-task experiments show that one fixed-capacity UHN can achieve competitive performance across diverse model families and datasets.
This suggests that the space of models and tasks considered here admits substantial shared regularities at the parameter level.
We also observe that removing the task-structure encoder often preserves final performance while slowing optimization, which is consistent with the interpretation that much of the transfer occurs through the shared parameter generator, whereas the task-structure encoder mainly improves conditioning and optimization dynamics.

\paragraph{Initialization and optimization stability.}
Initialization is important for stabilizing and accelerating training.
When the initialization phase is removed, optimization becomes slower and performance typically degrades in multi-model and multi-task settings.
In the recursive setting, training can diverge without initialization: errors in weight scale at higher recursion levels can compound through the generation chain, leading to exploding activations and unstable gradients at the leaf model.

\subsection{Limitations}
\paragraph{Potential performance gap versus direct training.}
UHN does not consistently match direct-training baselines on all tasks and architectures.

\paragraph{Limited regularization study.}
We do not systematically explore regularization strategies for hypernetwork training (e.g., weight decay \cite{loshchilov2017decoupled} or other priors), which may reduce overfitting and close performance gaps.

\paragraph{Incomplete support for non-trainable parameters.}
UHN does not generate non-trainable states such as BatchNorm \cite{ioffe2015batch} running statistics, which limits applicability to some architectures.

\paragraph{Compute overhead.}
Training a hypernetwork-to-network mapping can introduce additional optimization cost and memory usage, and it may require longer training schedules than direct training.

\subsection{Future Work}
\paragraph{Larger models and deeper recursion.}
Extending UHN to higher-capacity base models and deeper recursive model trees is a natural next step toward practical deployment.

\paragraph{Regularization for hypernetworks.}
A systematic study of regularization (e.g., weight decay \cite{loshchilov2017decoupled}, spectral constraints \cite{yoshida2017spectral}, and noise injection \cite{neelakantan2015adding}) may improve generalization and stability.

\paragraph{More expressive encodings.}
Incorporating more expressive coordinate encodings (e.g., hash encodings \cite{muller2022instant}) may improve convergence speed and sample efficiency.

\paragraph{Improving model-family generalization.}
When training on small model sets, improving robustness to rare or extreme architectures remains an important direction.

\paragraph{Generating non-trainable state.}
Extending the generator to produce auxiliary state (e.g., BatchNorm \cite{ioffe2015batch} statistics) would broaden the set of architectures supported.

\paragraph{Ensembling and uncertainty.}
Because UHN can efficiently instantiate many models, combining generated models via ensembling \cite{deutsch2019generative} and studying uncertainty \cite{ratzlaff2019hypergan} are promising directions.

%%%%%%%%%%%%%%%%%%%%%%%%%%%%%%%%%%%%%%%%%%%%%%%%%%%%%%%%%%%%%%%%%%%%%%%%%%%%%%%
%%%%%%%%%%%%%%%%%%%%%%%%%%%%%%%%%%%%%%%%%%%%%%%%%%%%%%%%%%%%%%%%%%%%%%%%%%%%%%%

\end{document}